\documentclass[journal]{IEEEtran}
\PassOptionsToPackage{table}{xcolor}
\usepackage[utf8]{inputenc}

\usepackage{hyperref}
\usepackage{multirow}
\usepackage{orcidlink}
\usepackage{subfigure}
\usepackage[numbers,sort&compress]{natbib}
\usepackage{amsmath}
\usepackage{amsfonts,amssymb}
\usepackage[table]{xcolor}
\usepackage{makecell}
\usepackage{threeparttable}
\usepackage{graphicx}
\usepackage{dblfloatfix}
\usepackage{booktabs}
\usepackage{tabularx}
\graphicspath{ {./figures/}{./figures_tgrs/}{./} }
\usepackage{tikz}
\usetikzlibrary{shapes.geometric}
\usepackage{microtype}
\usepackage{pifont}    
\usepackage{booktabs}  
\usepackage{graphicx}  

\hypersetup{
    colorlinks=true,
    linkcolor=cyan,
    filecolor=blue,
    urlcolor=black,
    citecolor=green,
}

\title{Multi-task Disentangling Fusion Mamba-CNN for MODIS Spectral-Temporal Unmixing with Endmember Variability Modeling}

\author{Motasem Alkayid, Yimin Zhu, Naser El-Sheimy,~\IEEEmembership{Member,~IEEE}, Lincoln Linlin Xu,~\IEEEmembership{Member,~IEEE}
\thanks{This work was supported by the Natural Sciences and Engineering Research Council of Canada (NSERC) under Grant RGPIN-2019-06744.}
\thanks{Yimin Zhu, Naser El-Sheimy, and Lincoln Linlin Xu are all with the Department of Geomatics Engineering,
University of Calgary, Canada (email: \{yimin.zhu, elsheimy, lincoln.xu\}@ucalgary.ca) (Corresponding author: Lincoln Linlin Xu).}
\thanks{Motasem Alkayid is with the Department of Geomatics Engineering, University of Calgary, Canada, and also with the Department of Geography, Faculty of Arts, The University of Jordan, Amman, Jordan (email: motasem.alkayid@ucalgary.ca)}
}

\begin{document}

\maketitle

\begin{abstract}
Spectral-temporal unmixing of MODIS time series is essential for applications such as abundance mapping and spectral library construction, yet it remains challenging due to (i) spectral-temporal information coupling in MODIS data, (ii) intra-class endmember spectral variability and inter-class signature ambiguity, and (iii) the instability of naively fusing multiple expert representations.
This paper presents the Multi-task Disentangling Fusion Mamba (MDF-Mamba) to address these limitations.
To resolve spectral-temporal coupling, we design a disentangling architecture that pairs per-channel spectral experts with a dedicated bidirectional state-space (Mamba) temporal branch. Convolutional modules each process a single spectral channel, while the Mamba branch encodes the cross-channel seasonal trajectory, forming a multi-branch expert ensemble with soft inter-expert reweighting based on squeeze-and-excitation.
For the integration of these spectral and temporal expert outputs, we introduce mHCFusion, adapting the manifold-constrained hyper-connection framework to merge the outputs in a stable and interpretable manner.
Endmember variability is addressed by extending the linear mixing model with pixel-wise learnable spectral variability weights that capture intra-class endmember variation.
Finally, we design a multi-task training objective that combines supervised abundance estimation with self-supervised spectral reconstruction and endmember regularization terms on the same training data, yielding a highly efficient framework for spectral-temporal unmixing.
Experiments on MODIS time series covering the Alberta and Saskatchewan provinces show that MDF-Mamba achieves the best performance on 12 of 14 evaluation metrics across both datasets (7 of 7 on Alberta; 5 of 7 on Saskatchewan), ranking second on the remaining two.
\end{abstract}


\section{Introduction}

\IEEEPARstart{E}{stimating} sub-pixel land cover fractions from moderate resolution satellite data is essential for broad-scale environmental observation. The Moderate Resolution Imaging Spectroradiometer (MODIS) provides daily global coverage at 250 - 1000\,m spatial resolution, producing dense spectral-temporal stacks that capture phenological dynamics across diverse biomes~\cite{ref_modis1, ref_modis2}. However, at this spatial resolution, most pixels contain mixtures of land cover types, necessitating spectral unmixing to estimate per-pixel fractional abundances. Accurate unmixing of MODIS time series is essential for applications such as crop monitoring, forest disturbance mapping, and land cover change detection across the heterogeneous landscapes of western Canada.

Traditional linear spectral unmixing (LSU) methods decompose each pixel spectrum as a linear combination of endmember signatures. On the other hand, fully Constrained Least Squares Unmixing (FCLSU)~\cite{ref_fclsu} solves this analytically via constrained least squares optimization~\cite{ref_lsu1}, but assumes fixed endmembers across the scene. Extensions such as sparse unmixing via variable splitting and augmented lagrangian (SUnSAL)~\cite{ref_sunsal} promote sparsity through $\ell_1$ regularization, while nonnegative matrix factorization (NMF) variants such as Spatial Group Sparsity NMF (SGSNMF)~\cite{ref_sgsnmf} incorporate spatial group sparsity via superpixel-guided regularization. 

While physically interpretable and computationally efficient, these approaches suffer from two fundamental limitations when applied to dense MODIS spectral-temporal stacks. Primarily, they process all spectral-temporal bands through a single shared model, conflating the physically distinct phenological cues of each spectral channel and preventing the disentanglement of channel-specific temporal dynamics. Additionally, they assume fixed endmember signatures across the entire scene, which is unrealistic given the substantial intra-class spectral variability driven by soil moisture, topography, and seasonal phenology~\cite{ref_emvar1, ref_emvar2}.

Deep learning approaches have emerged as viable alternatives for spectral unmixing tasks, offering the ability to learn nonlinear representations directly from data. Autoencoder-based methods such as PGMSU~\cite{ref_pgmsu} and SUnCNN~\cite{ref_suncnn} learn latent abundance representations through unsupervised reconstruction. Transformer-based architectures including DeepTrans-HSU~\cite{ref_deeptrans} and SRViT~\cite{ref_srvit} use self-attention to capture long-range spectral dependencies. Spatial-spectral attention networks such as SSAF-Net~\cite{ref_ssafnet} and PPM-Net~\cite{ref_ppmnet} jointly model spatial and spectral features through dual-stream encoders. More recently, 2024 frameworks have introduced contrastive learning~\cite{ref_contrastive_hsu}, continuous variational inference~\cite{ref_vi_unmixing}, and spectral variability-constrained diffusion models~\cite{ref_svcdm} to improve resilience against environmental noise and label scarcity. Despite these advances, deep learning methods largely inherit the same limitations: they process the full $L$-band stack through shared parameters, causing cross-channel interference; most retain fixed or weakly regularized endmembers that cannot capture intra-class spectral variability across the growing season; and when multi-stream features are aggregated, concatenation or pooling is used with no geometric guarantee on the stability or interpretability of the fusion.

The emergence of selective state space models (SSMs), particularly the Mamba architecture~\cite{ref_mamba}, offers a clear alternative for modeling long sequences with linear-time complexity $O(T)$ in sequence length. This makes the SSM well suited to the \emph{temporal} axis of MODIS time series, where land cover discrimination relies on the \emph{ordering} of phenological transitions (green-up, peak, senescence) rather than any single date spectrum~\cite{ref_tempcnn}. While 2024 has seen Mamba adapted for spatial-spectral hyperspectral classification~\cite{ref_s2mamba, ref_spectralmamba}, its application to unmixing remains underexplored. Recent work has applied Mamba to hyperspectral unmixing~\cite{ref_unmamba}, but scans the SSM over the spectral (band) dimension rather than time and does not address endmember variability. While Mamba has been applied to the temporal axis of satellite image time series for crop classification~\cite{ref_sitsmamba, ref_stsmamba}, a dedicated \emph{temporal} Mamba branch for phenology-driven \emph{unmixing} remains unexplored. More broadly, no existing method (traditional, deep learning, or SSM-based) jointly resolves all three limitations under a unified physics-consistent framework. Abundance supervision alone is insufficient to uniquely constrain the endmember geometry~\cite{ref_emvar1}: without a complementary spectral reconstruction objective, learned endmember embeddings can satisfy the abundance loss while converging to physically meaningless signatures. These open problems directly motivate the contributions of MDF-Mamba.

In this paper, we present the Multi-task Disentangling Fusion Mamba (MDF-Mamba) model, a novel architecture specifically designed for MODIS spectral-temporal unmixing. The key contributions of this work are summarized as follows:

\begin{enumerate}
    \item We design an asymmetric disentangling architecture via a hybrid CNN-Mamba multi-branch expert ensemble that assigns spectral-spatial encoding to fixed, channel-specialized convolutional experts and temporal sequence encoding to a dedicated Mamba branch, using complementary inductive biases. Unlike canonical mixture-of-experts (MoE), expert assignment is fixed by channel rather than learned via a gating network, with only the downstream combination weights learned. A cross-band input stem first learns inter-channel correlations before strided indexing splits the input into six per-channel spectral streams. Six spectral BandBranches independently process a single spectral channel each (NDVI, EVI, Red, NIR, Blue, MIR) using a dual-path 2D CNN (\textit{ElementBranchCNN}, $d{=}128$) that treats the $T = 23$ temporal observations as input channels. Each branch runs two parallel depth-three sub-networks, a spectral path ($1\!\times\!1$ convolutions) and a spatial path ($3\!\times\!3$ convolutions), cross-refined via a sigmoid-gated twist gate. A seventh TemporalMambaBranch processes all 23 composites jointly as a sequence of 6-dimensional multi-spectral tokens via a bidirectional selective state space model (SSM). After cross-modal alignment, a cross-branch Squeeze-and-Excitation (SE) module enables inter-expert communication by computing soft attention weights across the six aligned spectral streams. To couple spectral and temporal representations, each spectral stream is modulated by the temporal output via an adaptive spatial-channel phenology gate that uses depthwise $3\!\times\!3$ convolution and sigmoid gating to produce spatially varying injection maps, replacing scalar gates with content-dependent modulation. The six modulated streams are then hierarchically grouped by physical domain: vegetation indices (NDVI + EVI), optical reflectance (Red + NIR + Blue), and moisture (MIR), reducing the fusion input from six streams to three semantically coherent groups.

    \item To integrate the three hierarchically grouped expert streams, we employ a manifold-constrained hyper-connection fusion module (mHCFusion) that operates in a $d_{\text{fuse}}{=}64$ representation space via per-stream alignment projections, applies doubly stochastic Sinkhorn-projected mixing across the grouped streams, and produces a unified feature map. A scene-adaptive Feature-wise Linear Modulation (FiLM) classification head then conditions the fused representation on global scene context. An adaptive average pooling path generates per-channel scale and shift parameters that modulate the fused features before final class projection, allowing the model to adaptively adjust abundance boundary sharpness based on overall scene composition.

    \item We address endmember variability through an extended linear mixing model (ELMM) in which each of the $P$ land cover classes maintains $R$ learnable spectral variants. Initialization follows a two-tier purity protocol: variants are seeded from high-purity training pixels (dominant class fraction $\geq 0.80$); for biomes where no such pixels exist, a geographic grid fallback (minimum purity $\geq 0.30$) provides spatially representative initialization. A lightweight pixel-wise variability head produces per-pixel convex mixing coefficients over the $R$ variants. To capture non-linear mixing effects that the base ELMM cannot represent, a cross-attention reconstruction module allows each pixel to attend to the full $P{\cdot}R$ endmember variant bank to produce an attention-based spectrum, which is blended with the ELMM reconstruction via a learnable sigmoid-gated mixing coefficient. A Laplacian spectral smoothness prior further regularizes the endmember curves to enforce physical realism.

    \item We design a multi-task supervised-unsupervised learning framework with curriculum training that jointly optimizes supervised abundance estimation, unsupervised spectral reconstruction, per-branch regularization, spectral correlation preservation, and endmember smoothness. A dedicated curriculum learning strategy up-weights reconstruction losses during early training to stabilize the endmember dictionary before shifting optimization power toward abundance refinement, combined with cosine-annealed endmember anchoring that progressively releases the model from its initialization.
\end{enumerate}

Extensive experiments on MODIS time series datasets covering Alberta and Saskatchewan demonstrate that MDF-Mamba achieves the best performance on 12 of 14 evaluation metrics across abundance estimation (MAE, RMSE), spectral reconstruction (MSE, SAD), and classification accuracy (OA, AA, Kappa), ranking second on the remaining two. This outperforms eleven established baselines, including three traditional methods (FCLSU, SGSNMF, SUnSAL) and eight deep learning methods (DeepTrans-HSU, PGMSU, PPM-Net, SSAF-Net, SUnCNN, MiSiCNet, SRViT, and UNMamba). The remainder of this paper is organized as follows. Section~II reviews related work on spectral unmixing, endmember variability modeling, MoE fusion, and state space models. Section~III presents the proposed MDF-Mamba architecture in detail. Section~IV describes the experimental setup, datasets, and quantitative results. Section~V concludes the paper.

\section{Related Work}

\subsection{Linear and Deep Learning-Based Spectral Unmixing}

Classical spectral unmixing formulates the observed pixel spectrum $\mathbf{y} \in \mathbb{R}^{L}$ as a linear combination of $P$ endmember spectra collected in the matrix $\mathbf{M} \in \mathbb{R}^{L \times P}$, weighted by their corresponding fractional abundances $\mathbf{a} \in \mathbb{R}^{P}$:
\begin{equation}
    \mathbf{y} = \mathbf{M}\mathbf{a} + \mathbf{n}, \quad \text{s.t.} \quad \mathbf{a} \succeq \mathbf{0}, \quad \mathbf{1}^{\top}\mathbf{a} = 1
    \label{eq:lmm}
\end{equation}
where $\mathbf{n}$ denotes the additive noise. While traditional methods like FCLSU~\cite{ref_fclsu} solve~\eqref{eq:lmm} analytically, they are limited by the assumption that endmembers remain fixed across the entire scene. To overcome this and accommodate large spectral libraries, sparse unmixing methods such as SUnSAL~\cite{ref_sunsal} reformulate~\eqref{eq:lmm} as an $\ell_1$-regularized optimization problem. Solved via the alternating direction method of multipliers (ADMM), SUnSAL explicitly promotes sparsity in the fractional abundances.

Furthermore, nonnegative matrix factorization (NMF)-based approaches typically estimate endmembers and abundances jointly. Notably, SGSNMF~\cite{ref_sgsnmf} extends standard NMF by incorporating a spatial group sparsity regularizer that leverages superpixel segmentation to enforce local homogeneity, yielding improved spatial coherence in the abundance maps. In our baseline implementation (Section~IV-E), however, the endmembers are held fixed after Vertex Component Analysis (VCA) initialization to avoid model instability on near-collinear MODIS signatures.

Recent deep learning methods have shifted toward learning the unmixing function end-to-end. Early approaches employed autoencoder architectures, such as PGMSU~\cite{ref_pgmsu}, which constrain the latent space to represent abundance maps while using the decoder as a learnable endmember matrix. To better exploit spatial correlations, convolutional approaches like SUnCNN~\cite{ref_suncnn} and MiSiCNet~\cite{ref_misicnet} introduced architectures inspired by deep image priors for blind unmixing. However, because standard convolutions struggle to capture global context, transformer-based methods such as DeepTrans-HSU~\cite{ref_deeptrans} and SRViT~\cite{ref_srvit} employ self-attention to capture long-range spectral dependencies. Building on this, recent dual-stream architectures like SSAF-Net~\cite{ref_ssafnet} and PPM-Net~\cite{ref_ppmnet} employ spatial-spectral attention networks to explicitly model the interplay between local spatial context and global spectral features.

Methodological evolution in deep unmixing has increasingly focused on mitigating the reliance on labeled ground truth. Approaches now include contrastive self-supervised learning~\cite{ref_contrastive_hsu}, which learns robust representations by maximizing similarity between augmented views of the same pixel, and spectral variability-constrained diffusion models~\cite{ref_svcdm}, which iteratively generate robust abundance priors. However, while these generative and self-supervised advances help address data scarcity, many existing architectures are not explicitly designed to disentangle the complex spectral-temporal information structure inherent in broadband satellite time series. Furthermore, they often fail to exploit phenological dynamics related to spectral channels, leaving a critical gap that motivates the first contribution of this work.

\subsection{Endmember Variability Modeling}

The assumption of fixed endmembers fails at regional-scale studies, where intra-class spectral variation is heavily driven by soil moisture, topography, and seasonal phenology~\cite{ref_emvar1, ref_emvar2}. Early methods addressed this variability using straightforward physical constraints. For instance, ELMM~\cite{ref_elmm} allows for pixel-wise scaling of endmember spectra to account for illumination changes, while bundle-based approaches maintain extensive libraries of spectral variants for each class. More recently, the focus has shifted toward complex, data-driven modeling. Concurrent preprints such as $\lambda$Split~\cite{ref_lambdasplit} employ physics-informed deep networks to learn conditional distributions over concentrations, and BLUTH~\cite{ref_bluth} utilizes hierarchical unmixing to reveal endmembers across varying levels of spectral contrast. Pushing this further, cutting-edge 2024 approaches have introduced continuous variational inference frameworks~\cite{ref_vi_unmixing} and spectral variability-constrained diffusion models~\cite{ref_svcdm} to probabilistically handle environmental noise. However, while these generative approaches are powerful, they are highly computationally intensive. Crucially, they do not jointly couple endmember variability with multi-task sequence reconstruction under a single, unified deterministic loss.

\subsection{Mixture of Experts for Remote Sensing}

While standard Mixture of Experts (MoE) architectures rely on learned gating functions to dynamically route input tokens to specialized sub-networks, MDF-Mamba employs a different approach. As detailed in Section III-A, our architecture utilizes a fixed, channel-determined assignment paired with learned soft reweighting. Because it lacks the dynamic routing characteristic of strict MoE frameworks, we formally define MDF-Mamba's design as a multi-branch expert ensemble.

Recent frameworks have increasingly relied on conditional routing to manage complex data structures. For example, UniRoute~\cite{ref_uniroute} applies conditional routing to disentangle local details from global context. In remote sensing, MixtureRS~\cite{ref_mixtuers} applies sparse MoE layers to handle confusing land cover categories by assigning specialized experts to specific spectral-spatial patterns. Similarly, MaMOL~\cite{ref_mamol} introduces a modality-aware MoE to manage missing or coupled data.

However, stably fusing the outputs of these multi-stream architectures remains a challenge. To address this, the manifold-constrained hyper-connection (mHC) framework~\cite{ref_mhc1, ref_mhc2} provides a principled aggregation method. It projects connection weight matrices onto the Birkhoff polytope of doubly-stochastic matrices using the Sinkhorn-Knopp algorithm. This constraint ensures equal total signal throughput across all streams and preserves the identity-mapping property, which is essential for stable training. Adapting this robust mHC framework to fit the asymmetric spectral-temporal structure of MDF-Mamba is the focus of Contribution 2.

\subsection{State Space Models and Mamba for Remote Sensing Temporal Sequences}

State Space Models (SSMs), particularly the Mamba architecture, enable linear-time $\mathcal{O}(T)$ sequence modeling through an input-dependent selection mechanism~\cite{ref_mamba}. Unlike Transformers that rely on quadratic self-attention ($\mathcal{O}(T^2)$), Mamba recurrently updates its hidden state via:
\begin{equation}
    \mathbf{h}_t = e^{\mathbf{A} \odot \boldsymbol{\Delta}_t} \mathbf{h}_{t-1} + \boldsymbol{\Delta}_t \mathbf{B}_t \mathbf{u}_t
\end{equation}
where the input-dependent discretization parameter $\boldsymbol{\Delta}_t$ dynamically controls how much past information is retained or forgotten at each timestep. 

This adaptive gating mechanism maps directly onto crop phenological transitions. During early green-up transitions, the model enforces strong retention (``remember'' gating) to propagate key onset signals into the peak summer growing season. Conversely, during harvest or senescence phases, selective decay (``forget'' gating) allows the network to reset accumulated biomass states. In contrast, standard 3D convolutions apply static kernel weights uniformly across all timesteps, inherently limiting their capacity to accommodate inter-pixel phenological heterogeneity.

The asymmetry of MODIS time series, where most phenological information is encoded in the ordering of spectral transitions rather than the absolute reflectance values~\cite{ref_tempcnn}, further motivates SSM over convolutional architectures for temporal encoding. SSMs preserve temporal causality and are theoretically equivalent to infinite-receptive-field filters~\cite{ref_mamba}, whereas 3D convolutions are limited to the kernel window.

In remote sensing, state space models have seen rapid adoption since 2024, particularly for high dimensional spatial spectral feature extraction in hyperspectral classification via architectures such as S$^2$Mamba~\cite{ref_s2mamba} and SpectralMamba~\cite{ref_spectralmamba}. For spectral unmixing specifically, UNMamba~\cite{ref_unmamba} applies Mamba to demonstrate competitive performance with significantly reduced computational overhead compared to transformer-based alternatives. However, UNMamba and its contemporaries apply the SSM scan over the spectral (band) dimension rather than the temporal dimension, treating each wavelength as a sequence element. Closer to our setting, recent Mamba models scan the \emph{temporal} axis of satellite image time series for crop classification~\cite{ref_sitsmamba, ref_stsmamba}, confirming the value of SSMs for phenological sequence modeling, though they target hard classification rather than sub-pixel unmixing with endmember variability. MDF-Mamba addresses this gap via a hybrid CNN-Mamba architecture that applies the SSM selectively to the temporal axis while handling spectral-spatial structure through per-channel CNN experts (Contribution~1), fused via mHCFusion (Contribution~2), as detailed in Section~III.

\section{Proposed MDF-Mamba Architecture}

\begin{figure*}[!tbp]
    \centering
    \includegraphics[width=\textwidth]{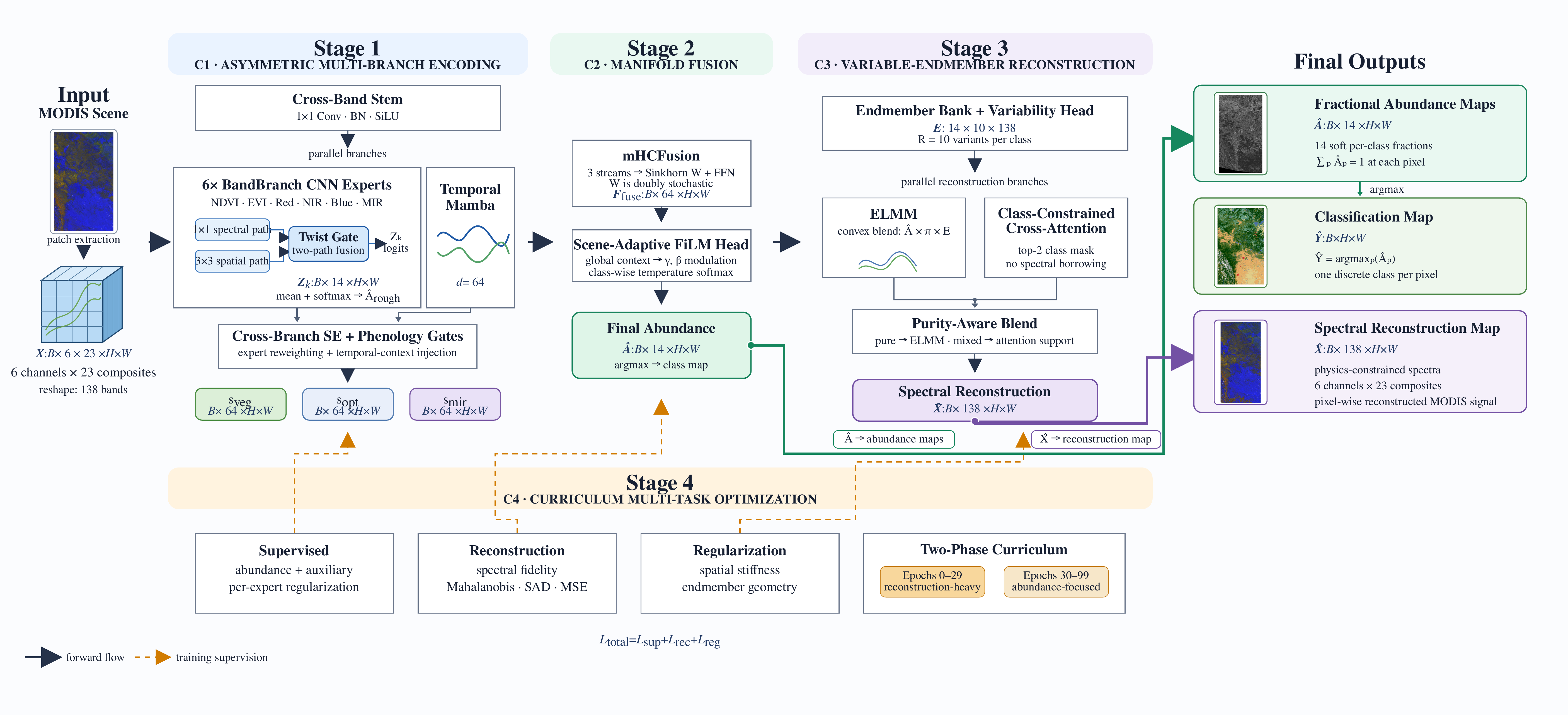}
    \caption{Overview of the proposed MDF-Mamba framework. The MODIS spectral--temporal input is processed through four stages. Stage~1 combines six band-specific CNN experts, each equipped with spectral and spatial paths and a Twist Gate, with a parallel Temporal Mamba branch. Cross-branch SE and phenology gates produce three complementary semantic streams. Stage~2 integrates these streams using manifold-constrained fusion and scene-adaptive modulation to estimate the fractional abundance map $\hat{\mathbf{A}}$. Stage~3 combines variable-endmember ELMM reconstruction and class-constrained cross-attention through a purity-aware blend to generate the spectral reconstruction $\hat{\mathbf{X}}$. The model produces fractional abundance maps, an abundance-derived classification map, and a physics-constrained spectral reconstruction map. Stage~4 jointly optimizes the abundance, reconstruction, and regularization objectives using a two-phase curriculum. Solid and dashed orange arrows represent forward flow and training supervision, respectively.}
    \label{fig:architecture}
\end{figure*}

Fig.~\ref{fig:architecture} illustrates the overall pipeline of the proposed MDF-Mamba model. Let $\mathbf{X} \in \mathbb{R}^{B \times L \times H \times W}$ denote a MODIS input patch, where $B$ is the batch size, $L = C \cdot T = 6 \times 23 = 138$ is the total number of spectral-temporal channels ($C = 6$ spectral channels, $T = 23$ time steps), and $H \times W$ is the spatial resolution. The model produces fractional abundance maps $\hat{\mathbf{A}} \in \mathbb{R}^{B \times P \times H \times W}$ for $P = 14$ land cover classes, along with a physics-constrained spectral reconstruction $\hat{\mathbf{X}} \in \mathbb{R}^{B \times L \times H \times W}$. The architecture is organized around four modules that correspond directly to the four contributions introduced in Section~I: Section~III-A presents the asymmetric multi-branch disentangling design; Section~III-B describes the manifold-constrained hyper-connection fusion; Section~III-C details the extended linear mixing model with pixel-wise endmember variability; and Section~III-D defines the multi-task supervised-unsupervised loss framework.

\subsection{Spectral-Temporal Disentangling via Multi-Branch Architecture}

The 138-band MODIS stack encodes $C = 6$ physically distinct spectral channels (NDVI, EVI, Red, NIR, Blue, MIR), each with its own phenological trajectory over $T = 23$ 16-day composites. Processing all $L$ channels through shared parameters conflates these cues. MDF-Mamba resolves this by assigning each spectral channel to a dedicated CNN expert, while a parallel Mamba branch encodes the temporal phenology shared across channels.

A shallow cross-band channel-mixing stem, implemented as a $1\!\times\!1$ convolution with batch normalization and SiLU activation, is first applied across all 138 bands to calibrate comparable feature scales and capture first-order inter-channel correlations (e.g., the Red--NIR relationship underlying vegetation indices); this brief joint step precedes the fully disentangled per-channel processing that follows, since all subsequent temporal-phenology encoding occurs independently within each of the six BandBranches, producing $\mathbf{X}' \in \mathbb{R}^{B \times L \times H \times W}$.

The output from this initial stem is then partitioned into six distinct per-channel spectral streams through strided channel indexing:
\begin{equation}
    \mathbf{X}_k = \mathbf{X}'[:, k::C, :, :] \in \mathbb{R}^{B \times T \times H \times W},\quad k = 0, \ldots, C{-}1
    \label{eq:band_split}
\end{equation}
Each stream is then processed by a dedicated BandBranch expert $f_k$ (\textit{ElementBranchCNN}, $d{=}128$). The expert works in three steps: two parallel feature paths, a \textit{twist gate} that lets the two paths exchange information, and a head that outputs class logits.

\textit{Parallel paths.} The spectral path uses only $1\!\times\!1$ convolutions, so it transforms each pixel's 23-step temporal profile independently of its neighbours. The spatial path uses $3\!\times\!3$ convolutions, so it also captures local spatial context such as field and patch boundaries. Each path stacks three blocks of convolution, batch normalization (BN), and GELU activation:
\begin{equation}
\begin{split}
    \boldsymbol{\Phi}^{\text{spec}}_k &= \bigl(\mathrm{GELU} \circ \mathrm{BN} \circ \mathrm{Conv}_{1\times1}\bigr)^{3}(\mathbf{X}_k), \\
    \boldsymbol{\Phi}^{\text{spat}}_k &= \bigl(\mathrm{GELU} \circ \mathrm{BN} \circ \mathrm{Conv}_{3\times3}\bigr)^{3}(\mathbf{X}_k),
\end{split}
    \label{eq:branch_paths}
\end{equation}
where $\circ$ denotes function composition, the exponent $3$ denotes three stacked blocks with separate weights, and both outputs lie in $\mathbb{R}^{B \times d \times H \times W}$.

\textit{Twist gate.} After the third block, each path is refined with information from the other path. Two gates are computed from the concatenated features of both paths, each with its own $1\!\times\!1$ convolution followed by a sigmoid $\sigma$:
\begin{equation}
\begin{split}
    \mathbf{G}^{\text{spat}}_k &= \sigma\!\left(\mathrm{Conv}_{1\times1}\!\left([\boldsymbol{\Phi}^{\text{spat}}_k \,\|\, \boldsymbol{\Phi}^{\text{spec}}_k]\right)\right), \\
    \mathbf{G}^{\text{spec}}_k &= \sigma\!\left(\mathrm{Conv}_{1\times1}\!\left([\boldsymbol{\Phi}^{\text{spat}}_k \,\|\, \boldsymbol{\Phi}^{\text{spec}}_k]\right)\right),
\end{split}
    \label{eq:twist_gate}
\end{equation}
where $\|$ denotes channel concatenation. Each gate takes a value in $(0,1)$ for every pixel and channel. $\mathbf{G}^{\text{spat}}_k$ controls how much spectral information is added to the spatial features, and $\mathbf{G}^{\text{spec}}_k$ controls how much spatial information is added to the spectral features:
\begin{equation}
\begin{split}
    \tilde{\boldsymbol{\Phi}}^{\text{spat}}_k &= \boldsymbol{\Phi}^{\text{spat}}_k + \mathbf{G}^{\text{spat}}_k \odot \boldsymbol{\Phi}^{\text{spec}}_k, \\
    \tilde{\boldsymbol{\Phi}}^{\text{spec}}_k &= \boldsymbol{\Phi}^{\text{spec}}_k + \mathbf{G}^{\text{spec}}_k \odot \boldsymbol{\Phi}^{\text{spat}}_k,
\end{split}
    \label{eq:twist_update}
\end{equation}

where $\odot$ denotes element-wise multiplication. The spatial features thus become aware of each pixel's spectral-temporal identity, and the spectral features become aware of local boundary geometry. Because the exchanged information is added to each path rather than replacing it, each path keeps its own features: a gate value near zero leaves the path unchanged, and a value near one adds the full features of the other path.

\textit{Classification head.} Finally, the two refined feature maps are joined along the channel dimension and passed to a small classification head $\psi_k$. The head first mixes the spatial and spectral features with a $1\!\times\!1$ convolution, BN, and GELU. A second $1\!\times\!1$ convolution then reduces them to $P$ channels, giving one score per land cover class at every pixel. These scores are the class logits of expert $k$, which are later turned into abundances by a softmax (Eq.~\eqref{eq:rough_abu}):

\begin{equation}
\begin{split}
    \mathbf{Z}_k = f_k(\mathbf{X}_k) &= \psi_k\!\left([\tilde{\boldsymbol{\Phi}}^{\text{spat}}_k \,\|\, \tilde{\boldsymbol{\Phi}}^{\text{spec}}_k]\right) \\
    &\in \mathbb{R}^{B \times P \times H \times W}, \quad k = 0, \ldots, C{-}1
\end{split}
    \label{eq:branch_enc}
\end{equation}

The six BandBranches form a fixed-assignment multi-branch expert ensemble, each specialising in one band and eliminating cross-band interference. Unlike conventional MoE, expert assignment is fixed by channel rather than learned via a gating network, with only the downstream combination weights learned (Cross-Branch SE, below). This design is motivated by the distinct per-channel temporal signatures in Fig.~\ref{fig:branch_grid}.

To encode cross-channel phenology, a seventh temporal branch utilizes a bidirectional selective state space model (SSM)~\cite{ref_sitsmamba, ref_stsmamba}. The 138-band stack is spatially downsampled $2\times$ and reshaped into per-pixel sequences of $T{=}23$ six-dimensional tokens, each linearly projected to $d_{\mathrm{model}}{=}64$. Each SSM layer performs input-dependent state transitions:
\begin{align}
    \mathbf{h}_t &= e^{\mathbf{A} \odot \boldsymbol{\Delta}_t} \odot \mathbf{h}_{t-1} + \boldsymbol{\Delta}_t \otimes \mathbf{B}_t \otimes \mathbf{U}_t \label{eq:ssm_h}\\
    \mathbf{y}_t &= \mathbf{C}_t^{\top} \mathbf{h}_t + \mathbf{D} \odot \mathbf{U}_t \label{eq:ssm_y}
\end{align}
where $\mathbf{A}$ is a learned decay matrix and the input-dependent $\boldsymbol{\Delta}_t$ is designed to control temporal memory length, shortening context during rapid phenological transitions and lengthening it during stable mid-summer periods. The bidirectional output is pooled by concatenating the annual mean with the first ($t{=}1$, January) and last ($t{=}T$, December) composite tokens, followed by a linear projection to reduce the concatenated $3d_{\mathrm{model}}{=}192$ representation down to $d_{\text{fuse}}{=}64$, then bilinearly upsampled to $(H,W)$ to give the temporal feature $\mathbf{a}_{\text{temp}}$. This $2\times$ downsample-then-upsample cycle trades fine spatial resolution in $\mathbf{a}_{\text{temp}}$ for tractable SSM compute.

The six spectral features and the temporal feature are projected to a shared $d_{\text{fuse}}{=}64$ space where a cross-branch squeeze-and-excitation module allows the experts to communicate through soft inter-expert attention weights. Each aligned spectral stream $\mathbf{a}'_k$ is modulated by the temporal feature via a spatially adaptive phenology gate:
\begin{equation}
    \tilde{\mathbf{a}}_k = \mathbf{a}'_k + \mathbf{b}_k \odot \mathbf{g}_k \odot \mathbf{a}_{\text{temp}}, \quad k = 0, \ldots, C{-}1
    \label{eq:pheno_gate}
\end{equation}
where $\mathbf{g}_k = \sigma\bigl(\mathrm{Conv}_{1\times1}([\mathbf{a}'_k \,\|\, \mathbf{a}_{\text{temp}}])\bigr)$ is a content-dependent gate and $\mathbf{b}_k = \sigma\bigl(\mathrm{BN}(\mathrm{DWConv}_{3\times3}(\mathbf{a}'_k))\bigr) \in (0,1)^{d_{\text{fuse}} \times H \times W}$ is a budget map learned from the spectral stream itself, with separate weights for each branch, so that pixels in transitional zones can receive stronger temporal context than homogeneous interiors.

Following modulation, the six streams are merged into three semantically coherent hierarchical groups by physical domain:
\begin{equation}
\begin{split}
    \mathbf{s}_{\text{veg}} &= f_{\text{veg}}([\tilde{\mathbf{a}}_0 \| \tilde{\mathbf{a}}_1]) \quad \text{(NDVI + EVI)}\\
    \mathbf{s}_{\text{opt}} &= f_{\text{opt}}([\tilde{\mathbf{a}}_2 \| \tilde{\mathbf{a}}_3 \| \tilde{\mathbf{a}}_4]) \quad \text{(Red + NIR + Blue)}\\
    \mathbf{s}_{\text{mir}} &= \tilde{\mathbf{a}}_5 \quad \text{(MIR)}
\end{split}
    \label{eq:hier_group}
\end{equation}
where $f_{\text{veg}}$ and $f_{\text{opt}}$ are $1\!\times\!1$ Conv + BN + SiLU projections, reducing the fusion input from six streams to $N{=}3$ and improving semantic composition.

This hierarchical grouping is physically motivated~\cite{ref_jensen2007}. By explicitly separating derived photosynthetic indices (Vegetation), structural surface reflectance (Optical), and water absorption properties (Moisture) into coherent physical domains, we prevent the fusion module from struggling to disentangle conflicting gradients and enforce a semantically structured fusion process.

A preliminary rough abundance map is obtained by averaging the six per-branch logits:
\begin{equation}
    \hat{\mathbf{A}}_{\text{rough}} = \text{softmax}\!\left(\frac{1}{C}\sum_{k=0}^{C-1} \mathbf{Z}_k\right) \in \mathbb{R}^{B \times P \times H \times W}
    \label{eq:rough_abu}
\end{equation}
with per-branch outputs $\hat{\mathbf{A}}_k = \text{softmax}(\mathbf{Z}_k)$ providing independent per-expert supervision targets (Section~III-D).

\subsection{Manifold-Constrained Hyper-Connection Fusion}

The $N{=}3$ semantic streams $\{\mathbf{s}_{\text{veg}}, \mathbf{s}_{\text{opt}}, \mathbf{s}_{\text{mir}}\}$ are integrated by the mHCFusion module~\cite{ref_mhc1, ref_mhc2} in a $d_{\text{fuse}}{=}64$ space. The streams are stacked and passed through two mHC blocks, each applying a \textit{width step} (Sinkhorn-projected doubly-stochastic mixing across streams, ensuring every stream sends and receives equal total weight without collapsing) and a \textit{depth step} (FFN with $4\times$ expansion). A softmax stream gate with RMSNorm and a residual skip then aggregate the streams into a fused map $\mathbf{F}_{\text{fuse}} \in \mathbb{R}^{B \times d_{\text{fuse}} \times H \times W}$.

A Feature-wise Linear Modulation (FiLM) classification head conditions the output on global scene context. Global average pooling (GAP) of $\mathbf{F}_{\text{fuse}}$ summarizes the scene, and a small MLP turns this summary into per-channel scale and shift parameters $\boldsymbol{\gamma}, \boldsymbol{\beta} \in \mathbb{R}^{d_{\text{fuse}}}$, which modulate the fused feature:
\begin{equation}
\begin{split}
    [\boldsymbol{\gamma} \,\|\, \boldsymbol{\beta}] &= \mathrm{MLP}\bigl(\mathrm{GAP}(\mathbf{F}_{\text{fuse}})\bigr), \\
    \mathbf{F}_{\text{mod}} &= \mathbf{F}_{\text{fuse}} \odot (1 + \boldsymbol{\gamma}) + \boldsymbol{\beta}
\end{split}
    \label{eq:film}
\end{equation}
The modulated feature $\mathbf{F}_{\text{mod}}$ is projected to $P$ class logits $\mathbf{F}$, and each logit is divided by a learnable per-class temperature $\tau_p \in [0.3, 2.0]$ before the softmax; a smaller $\tau_p$ gives sharper, more pure-like abundances:

\begin{equation}
    \hat{\mathbf{A}} = \text{softmax}\!\left(\mathbf{F} \oslash \boldsymbol{\tau}\right) \in \mathbb{R}^{B \times P \times H \times W}
    \label{eq:refined_abu}
\end{equation}
By applying the softmax activation across the class dimension, this formulation intrinsically enforces both the Abundance Sum-to-One Constraint (ASC, $\sum_p \hat{A}_p = 1$) and the Abundance Non-negativity Constraint (ANC, $\hat{A}_p \ge 0$) at each pixel, ensuring physically valid mixing fractions. Subsequently, the discrete land cover map for hard-classification evaluation is obtained via a pixel-wise argmax over the class dimension of $\hat{\mathbf{A}}$.
\vspace{2mm}

A single fixed endmember per class cannot represent intra-class spectral variability driven by soil moisture, topography, and phenology. MDF-Mamba adopts an Extended Linear Mixing Model (ELMM)~\cite{ref_elmm} with $R{=}10$ spectral variants per class.

The endmember variant bank $\mathbf{E} \in \mathbb{R}^{P \times R \times L}$ is derived from a learnable embedding matrix $\mathbf{Q} \in \mathbb{R}^{(P \cdot R) \times L}$ passed through a ReLU activation for non-negativity. The representative class mean for class $p$ is defined as:
\begin{equation}
    \bar{\mathbf{e}}_p = \frac{1}{R} \sum_{r=1}^{R} \mathbf{E}_{p, r, :}
\end{equation}
The rows of $\mathbf{Q}$ are initialized from high-purity training pixels with dominant abundance $\ge 0.80$ (supplemented by a $4 \times 5$ spatial grid fallback thresholded at $\ge 0.30$) and are constrained to the valid spectral range $[0, 1]$ throughout training.

A lightweight $1\times 1$ convolution $h_{\text{var}}$ ($P \to P R$ channels) acts as a variability head that maps the class logits $\mathbf{F}$ of Eq.~\eqref{eq:refined_abu}, taken before temperature scaling, to per-pixel variant mixing weights over the endmember bundles~\citep{somers2012automated}:
\begin{equation}
    \mathbf{W} = \text{softmax}_R\!\left(h_{\text{var}}(\mathbf{F})\right) \in \mathbb{R}^{B \times P \times R \times H \times W}
    \label{eq:var_weights}
\end{equation}
and the Extended Linear Mixing Model (ELMM)~\citep{drumetz2016blind} reconstruction combines class abundances with these pixel-adaptive convex combinations of variants:
\begin{equation}
    \hat{X}^{\text{ELMM}}_{b,l,h,w} = \sum_{p=0}^{P-1} \hat{A}_{b,p,h,w} \sum_{r=0}^{R-1} W_{b,p,r,h,w} \cdot E_{p,r,l}
    \label{eq:recon}
\end{equation}
While physically valid strictly for linear reflectance bands (e.g., Red, NIR, Blue, MIR), the model applies this linear mixing across all $L=138$ channels, including non-linear vegetation indices (NDVI, EVI). In line with feature-level self-supervised reconstruction objectives, this treats index mixing as a simplified approximation for deep representation learning rather than a strict physical radiative transfer law~\citep{settle1993linear}.

To model complex non-linear scattering effects, a lightweight cross-attention module is introduced to generate a non-linear feature reconstruction, $\hat{\mathbf{X}}_{\text{attn}}$. To mitigate spectral borrowing from erroneous endmember classes, this attention mechanism is constrained such that each pixel selectively attends only to the spectral variant banks associated with its top-two predicted classes.

The resulting non-linear representation is subsequently fused with the physics-driven ELMM reconstruction, $\hat{\mathbf{X}}_{\text{ELMM}}$, via a purity gating mechanism. This gate, defined as $\rho = \sigma\!\left((\max_p \hat{A}_p - 0.80) \times 10\right)$, adaptively quantifies pixel purity based on the maximum predicted fractional abundance. As formulated in Equation~\ref{eq:recon_blend}, $\rho$ acts as a soft routing function: highly pure pixels ($\rho \to 1$) are predominantly guided toward the physically constrained ELMM output. Conversely, for highly mixed pixels ($\rho \to 0$), the architecture adaptively incorporates the non-linear attention output. This integration is modulated by a learned, band-specific scaling parameter, $\sigma(\lambda_b)$ (empirically initialized to $0.2$), ensuring that non-linear adjustments do not induce spectral over-distortion:
\begin{equation}
    \hat{\mathbf{X}} = \bigl(1 - \sigma(\lambda_b)(1{-}\rho)\bigr)\hat{\mathbf{X}}_{\text{ELMM}} + \sigma(\lambda_b)(1{-}\rho)\hat{\mathbf{X}}_{\text{attn}}
    \label{eq:recon_blend}
\end{equation}
Because the cross-attention reconstructor can partially reproduce the input spectrum independently of the ELMM abundances, a low blended reconstruction error ($\hat{\mathbf{X}}$) evaluates the model's overall representation quality rather than the physical ELMM alone. We report the blended reconstruction metrics as a proxy for the overall framework's fidelity, while noting that pure ELMM reconstruction would be required to independently validate the abundance solution.

Where $\lambda_b$ is a learnable mixing scalar. To suppress high-frequency noise while preserving physically meaningful temporal continuity (e.g., phenological trajectories), a Laplacian smoothness penalty is applied to the mean endmembers, $\bar{\mathbf{E}}$. This penalty acts independently along the temporal dimension ($t \in [0, T-1]$) within each of the $C$ physical channels, thereby strictly avoiding artificial smoothing across distinct spectral channel boundaries:
\begin{equation}
\begin{split}
    \mathcal{L}_{\text{smooth}} &= \frac{1}{P \cdot C \cdot (T{-}2)} \sum_{p=0}^{P-1} \sum_{c=0}^{C-1} \sum_{t=1}^{T-2} \\
    &\quad \left(\bar{e}_{p,c,t+1} - 2\bar{e}_{p,c,t} + \bar{e}_{p,c,t-1}\right)^2
\end{split}
    \label{eq:loss_smooth}
\end{equation}

\subsection{Multi-Task Supervised-Unsupervised Learning}
\label{subsec:multitask}

MDF-Mamba is trained with a multi-task loss combining supervised abundance estimation, per-expert regularization, physics-constrained spectral reconstruction, spatial coherence regularization, and endmember regularization:
\begin{equation}
\begin{split}
    \mathcal{L} = \mathcal{L}_{\text{abu}} &+ 0.5\,\mathcal{L}_{\text{aux}} + 0.3\,\mathcal{L}_{\text{br}}\\
    + \text{boost}\cdot &\bigl(0.1\,\mathcal{L}_{\text{maha}} + 0.1\,\mathcal{L}_{\text{sad}} + 0.5\,\mathcal{L}_{\text{recon}} + 0.1\,\mathcal{L}_{\text{rsad}}\bigr)\\
    + \lambda_{\text{corr}}(t)\,\mathcal{L}_{\text{corr}} &+ 0.05\,\mathcal{L}_{\text{dir}} + \lambda_{\text{anc}}(t)\,\mathcal{L}_{\text{anc}} + \lambda_{\text{smooth}}(t)\,\mathcal{L}_{\text{smooth}}
\end{split}
    \label{eq:total_loss}
\end{equation}

The supervised abundance loss combines an L1 magnitude penalty with a KL shape penalty between the refined abundance $\hat{\mathbf{A}}$ and ground-truth labels $\mathbf{A}^*$:
\begin{equation}
    \mathcal{L}_{\text{abu}} = \|\hat{\mathbf{A}} - \mathbf{A}^*\|_1 + \lambda_{\text{kl}} \cdot \mathrm{KL}(\mathbf{A}^* \| \hat{\mathbf{A}})
    \label{eq:loss_abu}
\end{equation}
The same L1+KL form provides per-expert supervision on the rough abundance ($\mathcal{L}_{\text{aux}}$) and on each per-branch output ($\mathcal{L}_{\text{br}}$, averaged over $k$), so that every expert is supervised before phenology gating and no branch free-rides on the Mamba output. Five spectral reconstruction-related terms enforce fidelity between the blended reconstruction $\hat{\mathbf{X}}$ and the MODIS observation $\mathbf{X}$: a whitened Mahalanobis error ($\mathcal{L}_{\text{maha}}$), spectral angle distance ($\mathcal{L}_{\text{sad}}$, $\mathcal{L}_{\text{rsad}}$), and inverse-variance-weighted MSE ($\mathcal{L}_{\text{recon}}$) are jointly scaled by the curriculum \texttt{boost} factor, while a covariance-matrix MSE ($\mathcal{L}_{\text{corr}}$) preserving the inter-band correlation structure of MODIS data is scaled independently via $\lambda_{\text{corr}}(t)$. A spatial stiffness term $\mathcal{L}_{\text{dir}}$ penalises abundance discontinuities between spectrally similar neighbouring pixels (weighted by a Gaussian spectral-affinity kernel) to suppress salt-and-pepper noise. Finally, endmember regularisation keeps the learned signatures physically plausible: $\mathcal{L}_{\text{anc}}$ anchors the mean endmembers $\bar{\mathbf{E}}$ to their GT initialisation $\mathbf{E}^{(0)}$ with a weight $\lambda_{\text{anc}}(t)$ that is annealed over training, and $\mathcal{L}_{\text{smooth}}$ enforces spectral continuity as defined in~\eqref{eq:loss_smooth}. The loss weights follow a two-phase curriculum that emphasises reconstruction fidelity while the endmember dictionary stabilises and then shifts toward abundance accuracy; the exact schedule is given in Section~IV-C.

\section{Experiments}

\subsection{Datasets}

We evaluate the proposed MDF-Mamba model on a MODIS time series dataset covering the provinces of Alberta and Saskatchewan in western Canada. The dataset is constructed from the MOD13Q1 product (Collection~6.1), which provides 16-day composite vegetation indices and surface reflectance at 250\,m spatial resolution. The temporal stack spans the full 2010 calendar year (January--December) and includes six spectral channels: NDVI, EVI, Red, NIR, Blue, and MIR, each with 23 temporal observations, yielding $L = 138$ spectral-temporal bands per pixel.

Fractional abundance labels for the land cover classes provide the ground truth and are derived from the 2010 North American Land Change Monitoring System (NALCMS) at 30\,m resolution. The dataset comprises 14 distinct classes present within the Alberta study area. To maintain traceability with the broader 19-class North American spatial data framework, we retain the standardized integer class codes, which naturally omit tropical and absent sub-polar classes. The 14 evaluated classes are: (1)~Temperate/sub-polar needleleaf forest, (2)~Sub-polar taiga needleleaf forest, (5)~Temperate/sub-polar broadleaf deciduous forest, (6)~Mixed forest, (8)~Temperate/sub-polar shrubland, (10)~Temperate/sub-polar grassland, (11)~Sub-polar shrubland-lichen-moss, (12)~Sub-polar grassland-lichen-moss, (14)~Wetland, (15)~Cropland, (16)~Barren lands, (17)~Urban and built-up, (18)~Water, and (19)~Snow and ice.

Fractional abundances are computed by aggregating the 30\,m labels within each 250\,m MODIS pixel. An independent 250\,m land cover classification map is used as reference for pixel-level classification evaluation. Of these 14 classes, the two sub-polar lichen-moss classes (11, 12) are exceedingly rare and spectrally inseparable from background at MODIS resolution (Section~IV-B); they are therefore excluded from the evaluation, leaving $P = 12$ active classes for the final abundance and classification outputs. Note that the data-exploration figures (Figs.~\ref{fig:dataset_stats} and \ref{fig:branch_grid}) and spatial error distributions (Fig.~\ref{fig:spatial_error}b) retain 14 classes to visualize the lichen-moss classes' poor separability and their corresponding errors; the learned endmember analysis (Fig.~\ref{fig:endm_grid}) and the final global abundance and classification metrics (Tables~\ref{tab:main_results}--\ref{tab:sask_filtered}), however, use $P=12$, excluding both lichen-moss classes.

For data preparation, the study area is divided into non-overlapping patches of $64 \times 64$ pixels. Patches containing any no-data values are excluded. To ensure a rigorous evaluation of spatial generalization and prevent the spatial leakage inherent to random partitions~\cite{ref_roberts1997}, the dataset is split using a strict geographically-blocked hold-out protocol. The scene is partitioned into a regular grid of non-overlapping spatial blocks, with each block assigned exclusively to the training (70.0\%), validation (15\%), or test (15\%) partition. To explicitly prevent spatial autocorrelation, a strict exclusion buffer is enforced at every block boundary, guaranteeing that no $64 \times 64$ patch can straddle a split. Rather than extracting patches uniformly across the map, patches are drawn exclusively from the valid pools within their respectively assigned blocks, resulting in 2{,}731 training, 585 validation, and 585 test patches. All spectral bands are normalized to $[0, 1]$ using global min-max statistics computed from the training set.

\subsection{Data Exploration and Purity Analysis}

To better understand the dataset characteristics and justify the need for spectral unmixing, we performed an exploratory analysis on the extracted MODIS patches.

\begin{figure*}[!tbp]
\centering
\includegraphics[width=0.75\textwidth]{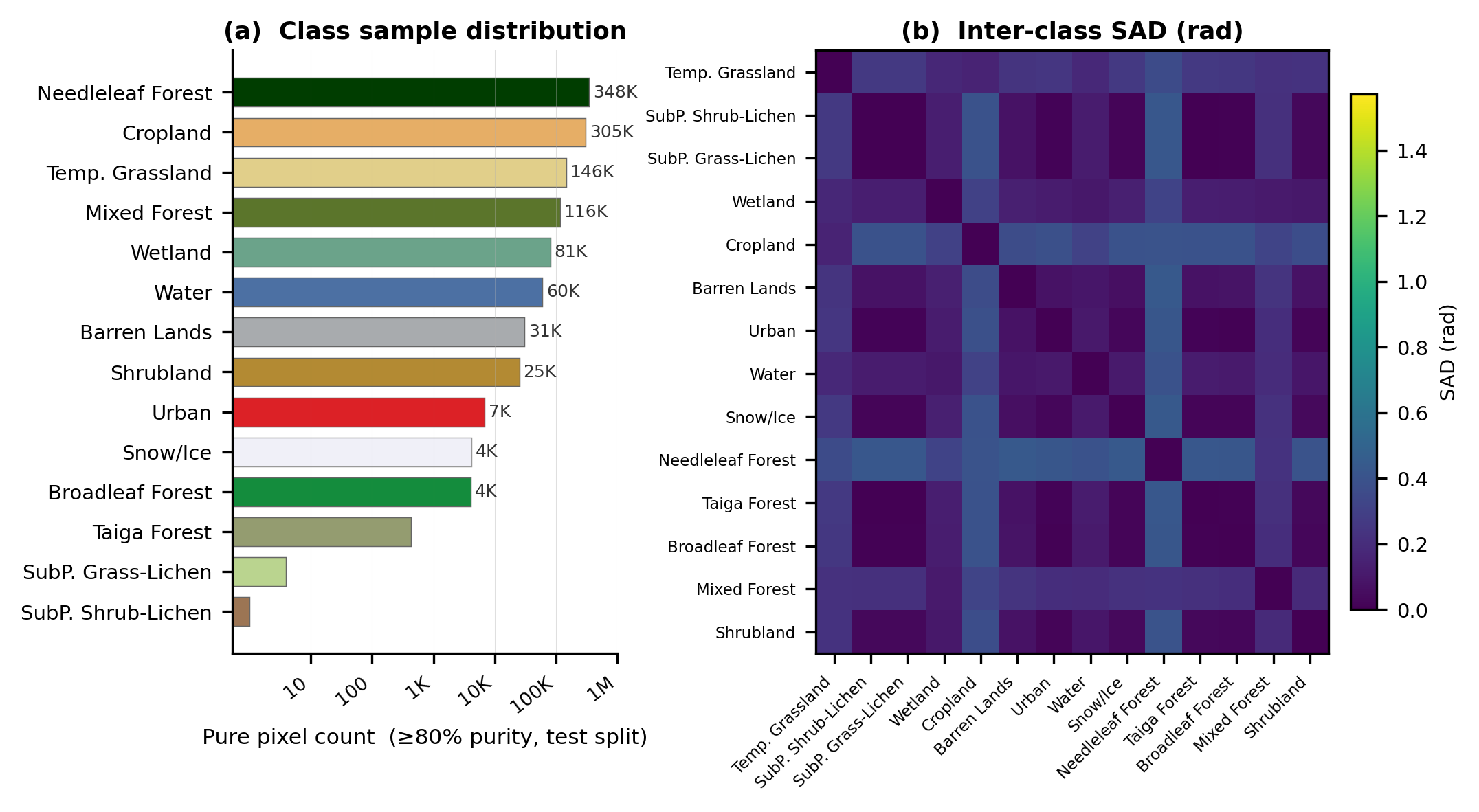}
\caption{Dataset statistics for the Alberta-Saskatchewan MODIS dataset. (a)~Pure-pixel sample counts per land cover class at 80\% abundance purity (log scale). (b)~Pairwise spectral angle distance (SAD) matrix between the 14 GT endmembers. Low SAD values ($<0.30$~rad, dark cells) identify class pairs that are spectrally confusable, motivating the multi-branch disentangling design.}
\label{fig:dataset_stats}
\end{figure*}

To evaluate class imbalance and inter-class separability, Fig.~\ref{fig:dataset_stats} summarises two complementary dataset properties. Panel~(a) shows the pure-pixel sample counts per class at 80\% abundance purity. The distribution spans nearly four orders of magnitude: Needleleaf Forest contributes over 3.8 million pure pixels while Urban and Broadleaf Forest each provide fewer than 400{,}000, creating a strongly imbalanced training signal that directly influences endmember quality. Notably, sub-polar lichen-moss classes (Sub-polar Shrub-Lichen and Sub-polar Grass-Lichen) each yield roughly 300{,}000 pixels yet exhibit near-zero spectral range across all 138 bands (range $< 0.08$ normalised units), indicating that even these nominally pure pixels are spectrally indistinguishable from mixed background. This observation motivates the geographic grid fallback initialisation of the endmember embedding described in Section~III-C.

Panel~(b) shows the pairwise spectral angle distance (SAD) matrix between the 14 GT endmembers computed from pure pixels. Several class pairs have low SAD ($< 0.30$~rad, corresponding to high spectral similarity): Sub-polar Shrub-Lichen vs.\ Sub-polar Grass-Lichen ($0.05$~rad), Needleleaf Forest vs.\ Mixed Forest ($0.18$~rad), and Wetland vs.\ Grassland ($0.22$~rad). These low-SAD pairs directly correspond to the classes with the highest per-class abundance MAE in Fig.~\ref{fig:spatial_error}(b) (needleleaf: 0.0801, mixed forest: 0.0574, wetland: 0.0569), consistent with inter-class spectral confusion being a major contributor to unmixing difficulty in these classes. We adopt $0.30$~rad (${\approx}17.2^\circ$) as the threshold for spectral confusability in the subsequent analysis; this boundary empirically captures the sharp inflection point in abundance MAE and aligns with standard angular tolerances for identifying highly correlated, structurally similar vegetation in multispectral classification~\cite{kruse1993spectral}.

\begin{figure}[!tbp]
\centering
\includegraphics[width=1\linewidth]{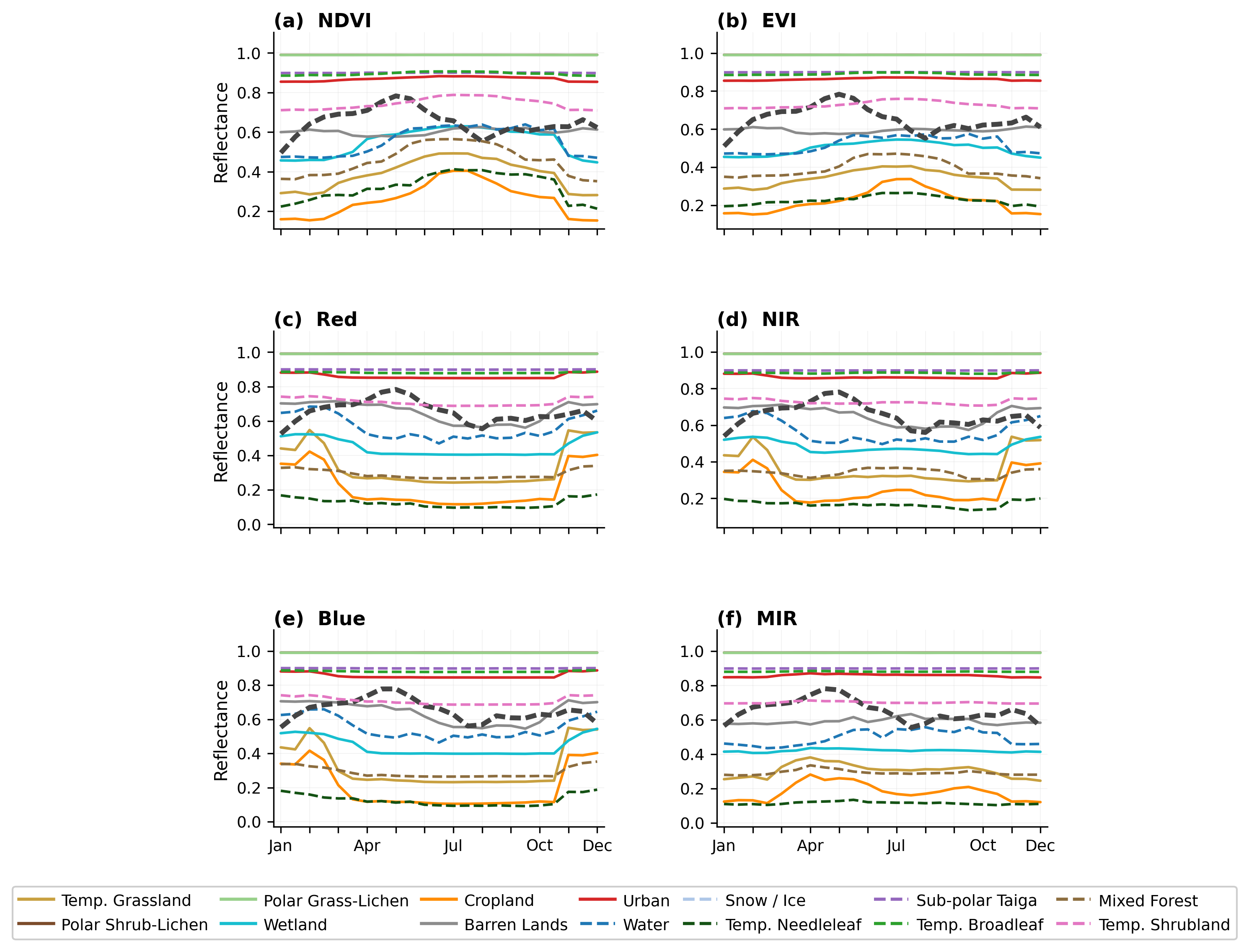}
\caption{Learned endmember temporal profiles (mean $\pm$1$\sigma$ over the $R{=}10$ variants) for 14 land cover classes across six spectral branches (NDVI, EVI, Red, NIR, Blue, MIR). Each panel shows the 23-step seasonal signature of every class as a distinct color-coded line with a faint $\pm$1 standard deviation band, illustrating which spectral dimensions provide the strongest class separability and motivating the multi-branch architecture.}
\label{fig:branch_grid}
\end{figure}

Analyzing branch-specific temporal signatures, Fig.~\ref{fig:branch_grid} presents the learned endmember temporal profiles for all 14 active land cover classes across the six spectral branches (NDVI, EVI, Red, NIR, Blue, MIR). Each panel shows one spectral branch with all 14 class signatures overlaid as color-coded lines over 23 MOD13Q1 composites spanning the full 2010 calendar year. The panels reveal markedly distinct inter-class separability per branch: NDVI captures the sharp summer green-up that separates cropland from evergreen forest classes~\cite{ref_huete2002}, NIR distinguishes coniferous from deciduous phenology via their late-season reflectance plateaus~\cite{ref_zhang2003pheno}, and MIR separates wetland from grassland through moisture-driven reflectance contrasts~\cite{ref_gao1996}. Critically, no single branch achieves full class separation (the lichen-moss biomes overlap with other classes in every branch), supporting the need for all six spectral dimensions and motivating the multi-branch disentangling architecture of MDF-Mamba.

\begin{figure*}[!tbp]
\centering
\includegraphics[width=0.85\textwidth]{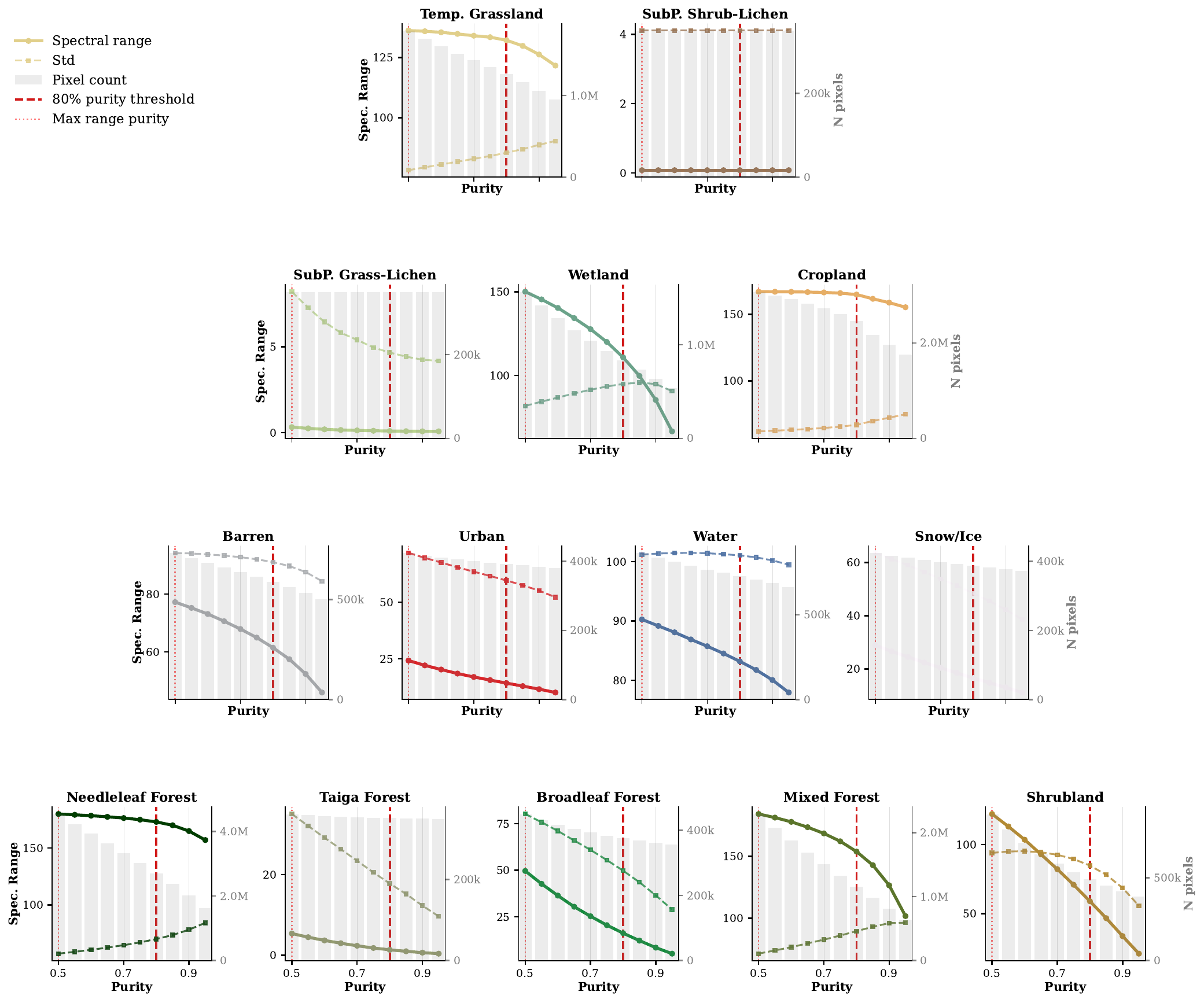}
\caption{Spectral range (solid line) and mean per-pixel standard deviation (dashed line) of each class as a function of the abundance purity threshold (0.50 - 0.95). Grey bars show the pixel count at each threshold (right axis). The red dashed vertical line marks the 80\% purity threshold used for GT endmember extraction. Classes whose spectral range remains near zero across all thresholds (Sub-polar Shrub-Lichen, Sub-polar Grass-Lichen) have no spectrally pure pixels at MODIS 250\,m resolution.}
\label{fig:purity_scan}
\end{figure*}

A purity analysis is conducted in Fig.~\ref{fig:purity_scan}, which presents the spectral range (max$-$min of the mean spectrum) and per-pixel standard deviation for each class as a function of the abundance purity threshold, swept from 0.50 to 0.95. The red dashed vertical line marks the 0.80 threshold used for GT endmember extraction in the proposed model. Two key observations emerge.

The pixel counts (grey bars) decrease steeply as the purity threshold rises from 0.50 to 0.95, confirming that highly pure pixels are rare at 250\,m MODIS resolution and that the vast majority of pixels contain contributions from multiple land cover classes. This prevalence of mixed pixels strictly necessitates sub-pixel spectral unmixing over hard-classification pipelines~\cite{ref_keshava2002, ref_bioucas2012}.

Additionally, the spectral characteristics behave differently across classes as purity increases. For spectrally rich vegetative classes such as Needleleaf Forest and Cropland, while the absolute spectral range slightly narrows, the mean standard deviation (variance) of the spectra actually rises or remains stable, indicating that higher purity pixels isolate a highly variable and representative phenological signature. In contrast, for sub-polar lichen-moss classes, the spectral range remains near zero across the entire threshold sweep, confirming the complete absence of spectrally distinguishable pure pixels for these biomes regardless of purity level. Similarly, non-vegetative classes such as Snow/Ice exhibit low and constant spectral variance because their reflectance lacks seasonal phenological dynamics. For intermediate classes such as Barren Lands, Urban, and Wetland, the spectral range peaks at lower purity thresholds and then decreases sharply, suggesting that the few very high purity pixels are unrepresentative outliers rather than canonical class signatures.

These observations directly informed three design decisions: (1)~the 80\% purity threshold for GT endmember extraction, which balances spectral representativeness against sample count for the majority of classes; (2)~the geographic fallback initialisation for classes where no high-purity pixels exist; and (3)~the $R = 10$ learnable spectral variants in the ELMM, which capture the intra-class variability that remains significant even at the maximum achievable purity.

\subsection{Implementation Details}

The model is implemented in PyTorch and trained using the Adam optimizer with an initial learning rate of $10^{-3}$ and a separate learning rate of $5 \times 10^{-4}$ for the endmember embedding $\mathbf{Q}$ and temperature $\boldsymbol{\tau}$ parameters. We employ a StepLR scheduler that decays the learning rate by a factor of 0.9 every 10 epochs. Training proceeds for 100 epochs with a batch size of 8 on a single NVIDIA A6000 GPU (48\,GB). Gradient clipping with a maximum norm of 1.0 is applied throughout training.

To solve the complex physical optimization landscape of multispectral unmixing, we employ a dedicated curriculum learning strategy. For the first 30 epochs, the model focuses heavily on physical consistency to stabilize the endmember dictionary by up-weighting all reconstruction-based losses by a factor of 3.0, strongly enforcing an endmember Laplacian smoothness prior ($\lambda_{\text{smooth}}=0.5$), and heavily weighting spectral band covariance ($\lambda_{\text{corr}}=1.0$). After epoch 30, the network relies on the converged dictionary and shifts its representational power toward abundance refinement by returning reconstruction penalties to their base weights ($\lambda_{\text{recon}} = 0.5$, $\lambda_{\text{sad}} = 0.1$). Concurrently, $\lambda_{\text{corr}}(t)$ anneals from $1.0$ to $0.2$, $\lambda_{\text{smooth}}(t)$ from $0.5$ to $0.05$, and the endmember anchor penalty $\lambda_{\text{anc}}(t)$ is cosine-annealed continuously from $1.0$ down to $0.1$ over the full 100 epochs. This ensures the model starts tightly anchored to the initial pure-pixel endmembers, but progressively gains the mathematical freedom to capture unseen non-linear variance via the adaptive cross-attention reconstructor.

The base loss weights are $\lambda_{\text{aux}} = 0.5$, $\lambda_{\text{br}} = 0.3$, $\lambda_{\text{maha}} = 0.1$, $\lambda_{\text{sad}} = 0.1$, $\lambda_{\text{rsad}} = 0.1$, $\lambda_{\text{dir}} = 0.05$, $\lambda_{\text{recon}} = 0.5$, $\lambda_{\text{corr}} = 0.2$, $\lambda_{\text{smooth}} = 0.05$. The number of endmember variants is $R = 10$. The hierarchical fusion module operates at $d_{\text{fuse}} = 64$ with $n_{\text{layers}} = 2$ inner mHC blocks on $N{=}3$ semantically grouped streams (vegetation, optical, moisture). The FiLM classification head uses a two-layer MLP ($64 \rightarrow 32 \rightarrow 128$) initialized to zero for near-identity early behaviour. The cross-attention reconstructor uses $d_{\text{attn}} = 64$ with the blend coefficient initialized at $\sigma(\lambda_b) \approx 0.2$. The endmember embedding $\mathbf{Q}$ is initialized from GT pure-pixel endmembers (purity $\geq 0.80$) with geographic fallback (purity $\geq 0.30$) as described in Section~III-C. The six \textit{ElementBranchCNN} BandBranches each use a hidden dimension of $d = 128$. The per-branch adaptive phenology gates use depthwise $3\!\times\!3$ convolution with sigmoid gating for spatially varying temporal injection. A Cross-Branch SE module provides inter-expert communication after alignment. The TemporalMambaBranch uses SSM hyperparameters $d_{\mathrm{model}} = 64$, $d_{\mathrm{state}} = 8$, and $n_{\mathrm{blocks}} = 2$ bidirectional SSM layers; the sequential for-loop scan is used for numerical stability in float32.

\subsection{Evaluation Metrics}

We evaluate performance across three complementary dimensions:

For abundance estimation, we compute the Sum of Absolute Errors (SAE, conventionally reported as MAE in some codebases but mathematically representing the sum of absolute errors over all $P$ classes bounded by 2.0) and Root Mean Square Error (RMSE) between predicted and ground-truth fractional abundances, averaged across all test pixels.

For spectral reconstruction, we calculate the Mean Squared Error (MSE) and Spectral Angle Distance (SAD, in radians) between the physics reconstructed spectrum $\hat{\mathbf{x}}$ and the observed MODIS input $\mathbf{x}$, measuring the physical consistency of the unmixing solution:
\begin{equation}
    \text{SAD}(\mathbf{x}, \hat{\mathbf{x}}) = \arccos\!\left(\frac{\mathbf{x}^{\top}\hat{\mathbf{x}}}{\|\mathbf{x}\|\,\|\hat{\mathbf{x}}\|}\right)
    \label{eq:sad}
\end{equation}

For classification evaluation, we report Overall Accuracy (OA), Average Accuracy (AA), and Cohen's Kappa coefficient ($\kappa$), computed by assigning each pixel to the class with the highest predicted abundance and comparing against the 250\,m land cover reference map. For detailed class-stratified analysis under imbalanced distributions, we additionally report the per-class F1 score, representing the balanced harmonic mean of precision and recall for individual land cover types.

\subsection{Baseline Methods}

We compare MDF-Mamba against eleven representative unmixing methods spanning traditional optimization, NMF-based, autoencoder-based, CNN-based, spatial-spectral attention-based, transformer-based, and SSM-based frameworks:

\begin{enumerate}
    \item \textbf{FCLSU}~\cite{ref_fclsu}: Fully Constrained Least Squares Unmixing, solved via the standard non-negative least-squares (NNLS) formulation with an appended sum-to-one row, using VCA-initialized endmembers. As a purely spectral method, FCLSU is inherently limited on MODIS broadband data: VCA-extracted endmembers for spectrally similar classes (e.g., different forest or shrubland types) exhibit pairwise cosine similarities exceeding 0.99, rendering the normal equations ill-conditioned and collapsing predictions onto the dominant classes.
    \item \textbf{FE-NMF} (Modified SGSNMF)~\cite{ref_sgsnmf}: A Fixed-Endmember Non-negative Matrix Factorization control baseline. In our implementation, endmembers are initialised from VCA and held \emph{fixed} during multiplicative abundance updates (only the abundance matrix is optimised); allowing joint endmember refinement with near-collinear MODIS signatures causes endmember collapse. Superpixel-guided spatial group sparsity from the canonical SGSNMF is omitted.
    \item \textbf{SUnSAL}~\cite{ref_sunsal}: Sparse Unmixing via variable Splitting and Augmented Lagrangian, an ADMM-based sparse unmixing method. While canonical SUnSAL is formulated with strict sum-to-one constraints, our evaluation tracks sparsity penalties independently of the strict sum constraint to avoid the condition where the $\ell_1$ penalty becomes a constant $1$.
    \item \textbf{DeepTrans-HSU}~\cite{ref_deeptrans}: A transformer-based autoencoder architecture with multi-head self-attention for spectral feature learning.
    \item \textbf{PGMSU}~\cite{ref_pgmsu}: A probabilistic generative model for spectral unmixing with Dirichlet-distributed abundances.
    \item \textbf{PPM-Net}~\cite{ref_ppmnet}: A pixel-level probabilistic mixture network with spatial-spectral attention and KL-divergence regularization.
    \item \textbf{SSAF-Net}~\cite{ref_ssafnet}: A spatial-spectral attention fusion network with dual-stream encoding and cross-attention feature integration.
    \item \textbf{SUnCNN}~\cite{ref_suncnn}: A deep image prior-inspired convolutional architecture for blind spectral unmixing.
    \item \textbf{MiSiCNet}~\cite{ref_misicnet}: A multi-scale convolutional network with skip connections and endmember estimation head.
    \item \textbf{SRViT}~\cite{ref_srvit}: A self-supervised relation-aware vision transformer with pixel-position coding for unmixing.
    \item \textbf{UNMamba}~\cite{ref_unmamba}: A state space model-based architecture for hyperspectral unmixing leveraging the Mamba selective scan mechanism.
\end{enumerate}

To ensure a fair comparison, all deep learning baselines are trained under an identical supervised protocol: the same training/validation/test splits, the same $64 \times 64$ patch extraction, the same VCA-initialized endmembers, and the same L1 abundance loss with ground-truth supervision. The three traditional baselines (FCLSU, FE-NMF, SUnSAL) are applied directly to the test set without training, all initialised from the same VCA endmembers. We note that the proposed MDF-Mamba uses GT-derived high purity pixels for its endmember initialization rather than unsupervised VCA, reflecting its supervised design. All models are evaluated using the shared \texttt{evaluate\_and\_save} utility to guarantee consistent metric computation.

\subsection{Quantitative Results}

Tables~\ref{tab:main_results} and~\ref{tab:sask_results} present the quantitative comparison of MDF-Mamba against eleven baseline methods on the Alberta and Saskatchewan test sets, respectively. Our model achieves the best or second-best performance across all evaluation metrics on both datasets, demonstrating consistent generalisation across geographically distinct MODIS landscapes.

\begin{table*}[!t]
\centering
\caption{Quantitative performance comparison on the Alberta MODIS dataset. Bold indicates the best result; underline indicates the second best. $\downarrow$ indicates lower is better; $\uparrow$ indicates higher is better.}
\label{tab:main_results}
\begin{threeparttable}
\resizebox{\textwidth}{!}{
\begin{tabular}{l l | c c | c c | c c c}
\toprule
\textbf{Type} & \textbf{Model} & \multicolumn{2}{c|}{\textbf{Abundance}} & \multicolumn{2}{c|}{\textbf{Reconstruction}} & \multicolumn{3}{c}{\textbf{Classification}} \\
 & & \textbf{MAE}$\downarrow$ & \textbf{RMSE}$\downarrow$ & \textbf{MSE}$\downarrow$ & \textbf{SAD (rad)}$\downarrow$ & \textbf{OA}$\uparrow$ & \textbf{AA}$\uparrow$ & $\boldsymbol{\kappa}$$\uparrow$ \\
\midrule
\multirow{3}{*}{Traditional}
& FCLSU~\cite{ref_fclsu}               & 0.0633 & 0.1530 & 0.0106 & 0.2445 & 0.5256 & 0.3290 & 0.4256 \\
& SGSNMF~\cite{ref_sgsnmf}             & 0.0703 & 0.1668 & 0.0118 & 0.2512 & 0.4734 & 0.3258 & 0.3750 \\
& SUnSAL~\cite{ref_sunsal}            & 0.0666 & 0.1547 & \underline{0.0104} & 0.2469 & 0.5124 & 0.2984 & 0.4035 \\
\midrule
\multirow{3}{*}{CNN/AE}
& PGMSU~\cite{ref_pgmsu}               & 0.0329 & 0.0842 & 0.0116 & 0.2344 & 0.6847 & 0.4152 & 0.6159 \\
& SUnCNN~\cite{ref_suncnn}             & 0.0309 & 0.0823 & 0.0141 & 0.2678 & \underline{0.6985} & 0.4308 & 0.6328 \\
& MiSiCNet~\cite{ref_misicnet}         & 0.0320 & 0.0828 & 0.0141 & 0.2695 & 0.6982 & \underline{0.4451} & \underline{0.6333} \\
\midrule
\multirow{2}{*}{CNN+Attention}
& SSAF-Net~\cite{ref_ssafnet}          & \underline{0.0301} & \underline{0.0755} & 0.0274 & 0.3646 & 0.6957 & 0.4450 & 0.6290 \\
& PPM-Net~\cite{ref_ppmnet}            & 0.0314 & 0.0792 & 0.0288 & 0.3530 & 0.6873 & 0.4296 & 0.6184 \\
\midrule
\multirow{2}{*}{Transformer}
& DeepTrans-HSU~\cite{ref_deeptrans}   & 0.0580 & 0.1362 & 0.0134 & 0.2551 & 0.5351 & 0.2473 & 0.4167 \\
& SRViT~\cite{ref_srvit}               & 0.0591 & 0.1363 & 0.0294 & 0.2977 & 0.5208 & 0.2269 & 0.4016 \\
\midrule
SSM
& UNMamba~\cite{ref_unmamba}             & 0.0366 & 0.0932 & 0.0105 & \underline{0.2232} & 0.6797 & 0.4084 & 0.6106 \\
\midrule
\rowcolor{yellow!15}
\textbf{Ours} & \textbf{MDF-Mamba}                    & \textbf{0.0255} & \textbf{0.0654} & \textbf{0.0067} & \textbf{0.1843} & \textbf{0.7122} & \textbf{0.4721} & \textbf{0.6575} \\
\bottomrule
\end{tabular}
}
\end{threeparttable}
\end{table*}

Several key observations can be drawn from Table~\ref{tab:main_results}:

Regarding abundance estimation, the proposed MDF-Mamba achieves the best MAE of 0.0255 and RMSE of 0.0654, outperforming all baselines including SSAF-Net (MAE: 0.0301, a 15.3\% reduction) and substantially surpassing all three traditional baselines - FCLSU (MAE: 0.0633, a 59.7\% reduction), SGSNMF, and SUnSAL - as well as DeepTrans-HSU (MAE: 0.0580, a 56.0\% reduction). The performance gap between traditional baselines and MDF-Mamba reflects a fundamental limitation of applying spectral-only LMM methods to MODIS broadband data: VCA-extracted endmembers for spectrally similar classes exhibit pairwise cosine similarities exceeding 0.99, rendering the unmixing problem ill-conditioned. FCLSU and SUnSAL both suffer from this ill-conditioning, causing predictions to collapse onto the spectrally dominant classes. SUnSAL additionally compounds this with its $\ell_1$ sparsity penalty, which by design activates only a few endmembers per pixel. Among the three traditional methods, SUnSAL achieves the lowest reconstruction MSE (0.0104); SGSNMF, with endmembers held fixed during NMF updates, yields a higher MSE of 0.0118, and its classification accuracy (OA: 0.4734) still falls substantially below all deep learning baselines. The proposed method overcomes these limitations by exploiting temporal phenology as a discriminative signal that separates classes indistinguishable in any single-timestep broadband spectrum, directly validating our first Contribution. This demonstrates the advantage of the multi-task learning framework combining temporal phenology encoding with physics-constrained reconstruction~\cite{ref_zhang2017mtl, ref_caruana1997}.

In terms of spectral reconstruction, MDF-Mamba achieves the lowest reconstruction SAD (0.1843 rad) and MSE (0.0067) among all evaluated methods, outperforming the second-best UNMamba (SAD: 0.2232 rad) by 17.4\% and the second-best SUnSAL (MSE: 0.0104) by 35.6\%. This dual leadership on both reconstruction metrics demonstrates that the multi-task physics-constrained loss (fourth contribution), combined with the ELMM endmember variability module (third contribution), enables MDF-Mamba to simultaneously preserve spectral \emph{shape} fidelity and per-band magnitude accuracy across the full 138-band temporal sequence. The SAD metric is particularly sensitive to spectral shape rather than magnitude~\cite{ref_kruse1993}, indicating that the TemporalMambaBranch's input-adaptive SSM encoding effectively captures the unique phenological signatures within each pixel.

For classification accuracy, MDF-Mamba achieves the best performance across all three classification metrics: overall accuracy (OA: 0.7122), average accuracy (AA: 0.4721), and Kappa ($\kappa$: 0.6575), outperforming SUnCNN (OA: 0.6985), MiSiCNet (AA: 0.4451, $\kappa$: 0.6333), and SSAF-Net (AA: 0.4450). While the absolute AA values remain relatively low across all models due to the extreme long tail class imbalance inherent to the study region, which severely penalizes minority classes, MDF-Mamba nonetheless leads on AA, the metric most sensitive to this minority class performance. This confirms that the mHC stream-mixing fusion with learnable stream gates (second contribution) successfully preserves per-class discriminative signals during multi-stream integration, enabling improved classification of spatially rare and spectrally ambiguous classes.

Table~\ref{tab:sask_results} reports results on the Saskatchewan test set, which represents a geographically distinct region with different land cover composition and seasonal dynamics compared to Alberta. The Saskatchewan landscape is dominated by agricultural croplands interspersed with wetlands, urban areas, and boreal forest patches~\cite{ref_aafc2010, ref_beaudoin2014}, creating denser spectral mixtures than the more heterogeneous Alberta terrain.

\begin{table*}[!t]
\centering
\caption{Quantitative performance comparison on the \textbf{Saskatchewan} MODIS dataset. Bold indicates the best result; underline indicates the second best. $\downarrow$ indicates lower is better; $\uparrow$ indicates higher is better.}
\label{tab:sask_results}
\begin{threeparttable}
\resizebox{\textwidth}{!}{
\begin{tabular}{l l | c c | c c | c c c}
\toprule
\textbf{Type} & \textbf{Model} & \multicolumn{2}{c|}{\textbf{Abundance}} & \multicolumn{2}{c|}{\textbf{Reconstruction}} & \multicolumn{3}{c}{\textbf{Classification}} \\
 & & \textbf{MAE}$\downarrow$ & \textbf{RMSE}$\downarrow$ & \textbf{MSE}$\downarrow$ & \textbf{SAD (rad)}$\downarrow$ & \textbf{OA}$\uparrow$ & \textbf{AA}$\uparrow$ & $\boldsymbol{\kappa}$$\uparrow$ \\
\midrule
\multirow{3}{*}{Traditional}
& FCLSU~\cite{ref_fclsu}               & 0.0729 & 0.2217 & 0.0126 & 0.2560 & 0.5572 & 0.2729 & 0.4495 \\
& SGSNMF~\cite{ref_sgsnmf}             & 0.0691 & 0.2481 & 0.0312 & 0.3498 & 0.5229 & 0.1857 & 0.3822 \\
& SUnSAL~\cite{ref_sunsal}            & 0.0807 & 0.2359 & 0.0132 & 0.2533 & 0.4607 & 0.2417 & 0.3476 \\
\midrule
\multirow{3}{*}{CNN/AE}
& PGMSU~\cite{ref_pgmsu}               & 0.0546 & 0.1748 & 0.0143 & 0.2617 & 0.6986 & 0.3832 & 0.6286 \\
& SUnCNN~\cite{ref_suncnn}             & \underline{0.0509} & 0.1685 & 0.0162 & 0.2836 & \underline{0.7176} & 0.3874 & 0.6493 \\
& MiSiCNet~\cite{ref_misicnet}         & 0.0532 & 0.1685 & 0.0164 & 0.2858 & 0.7121 & 0.4175 & 0.6450 \\
\midrule
\multirow{2}{*}{CNN+Attention}
& SSAF-Net~\cite{ref_ssafnet}          & 0.0544 & 0.1766 & \textbf{0.0063} & \textbf{0.1823} & 0.6829 & 0.3995 & 0.6116 \\
& PPM-Net~\cite{ref_ppmnet}            & 0.0542 & 0.1701 & 0.0226 & 0.3506 & 0.7064 & 0.4199 & 0.6377 \\
\midrule
\multirow{2}{*}{Transformer}
& DeepTrans-HSU~\cite{ref_deeptrans}   & 0.0753 & 0.2097 & 0.0162 & 0.2837 & 0.5564 & 0.2231 & 0.4346 \\
& SRViT~\cite{ref_srvit}               & 0.0760 & 0.2259 & 0.0198 & 0.3082 & 0.5197 & 0.2033 & 0.3850 \\
\midrule
SSM
& UNMamba~\cite{ref_unmamba}           & 0.0539 & \underline{0.1669} & 0.0133 & 0.2492 & 0.7167 & \underline{0.4330} & \underline{0.6499} \\
\midrule
\rowcolor{yellow!15}
\textbf{Ours} & \textbf{MDF-Mamba}     & \textbf{0.0492} & \textbf{0.1575} & \underline{0.0093} & \underline{0.2081} & \textbf{0.7485} & \textbf{0.4779} & \textbf{0.6881} \\
\bottomrule
\end{tabular}
}
\end{threeparttable}
\end{table*}

On Saskatchewan, MDF-Mamba achieves the best performance on five of the seven evaluation metrics, confirming strong cross-regional generalization without domain-specific retraining. It leads on all three classification metrics (OA: 0.7485, AA: 0.4779, $\kappa$: 0.6881), on abundance MAE (0.0492) and RMSE (0.1575), and ranks second on both reconstruction MSE (0.0093) and SAD ($0.2081$~rad) behind SSAF-Net (MSE: 0.0063, SAD: $0.1823$~rad). This highly competitive reconstruction performance validates the robustness of the ELMM-based endmember variability module and the multi-task learning framework under domain shift: the physics-constrained reconstruction head maintains physically meaningful spectral fidelity even when the endmember dictionary, learned on Alberta, encounters Saskatchewan's distinct cropland-dominated spectral mixtures~\cite{ref_willmott2005, ref_chai2014}.

For classification, MDF-Mamba achieves the best performance across all three metrics: OA (0.7485), AA (0.4779), and $\kappa$ (0.6881), outperforming all baselines including SUnCNN (OA: 0.7176), UNMamba (AA: 0.4330, $\kappa$: 0.6499), and MiSiCNet (AA: 0.4175). The fact that MDF-Mamba leads on AA - the metric most sensitive to minority class performance - validates Contribution~1 (asymmetric disentangling) and Contribution~2 (mHC fusion): the multi-branch architecture successfully transfers its per-channel phenological representations to a geographically distinct landscape without overfitting to Alberta-specific class distributions, while the manifold-constrained fusion preserves minority-class signals that single-stream architectures suppress. MDF-Mamba maintains a substantial lead over all traditional baselines ($>$34.3\% OA improvement) and the Transformer-based SRViT ($>$44.0\% OA improvement), demonstrating consistent superiority across geographically diverse MODIS landscapes.

The higher OA and $\kappa$ on Saskatchewan (0.7485 and 0.6881) than on Alberta (0.7122 and 0.6575) should not be read as better performance than within the training province. OA weights every pixel equally and is therefore dominated by the most frequent classes. Cropland and Water, the two classes that transfer most reliably (F1 of 0.934 and 0.816, Table~\ref{tab:cross_province}), together account for 38.6\% of the Saskatchewan abundance (26.3\% and 12.3\%), compared with 21.4\% in Alberta (17.2\% and 4.2\%). Conversely, Mixed Forest, one of the most challenging classes for abundance estimation, falls from 15.1\% of the Alberta abundance to 3.4\% in Saskatchewan. The Saskatchewan class composition therefore favors easily separable classes, which raises OA and $\kappa$ even though the abundance error increases (MAE: 0.0255 to 0.0492).

Saskatchewan is evaluated as a zero-shot cross-provincial transfer: the model is trained exclusively on Alberta and applied to the full Saskatchewan MODIS scene without retraining or fine-tuning, using the Alberta global band normalization statistics. The two provinces differ markedly in class composition: Cropland rises from 17.2\% of the Alberta training abundance to 26.3\% in Saskatchewan (40.0\% of its pure pixels), Water from 4.2\% to 12.3\%, while Mixed Forest falls from 15.1\% to 3.4\%. Full-scene maps are produced with non-overlapping $64 \times 64$ tiles; the main comparison (Table~\ref{tab:sask_results}) uses all valid pixels against the 250\,m reference map, with each pixel's reference abundance set to a one-hot vector of its mapped class, and the filtered comparison (Table~\ref{tab:sask_filtered}) uses every third valid patch, restricted to pixels with dominant-class purity $\ge 80\%$ and evaluated against the fractional abundances. The abundance errors in Table~\ref{tab:sask_results} are therefore not directly comparable with those in Table~\ref{tab:main_results}, which are computed against fractional references.

\subsection{Quantitative Results on Filtered High-Purity Datasets}

To further evaluate the robustness of the proposed model on highly confident ground-truth regions, we additionally test all methods on filtered subsets of the Alberta and Saskatchewan datasets containing only pixels with dominant class purity $\ge 80\%$. Tables~\ref{tab:alberta_filtered} and~\ref{tab:sask_filtered} present the quantitative comparison on these high-purity subsets.

\begin{table*}[!t]
\centering
\caption{Quantitative performance comparison on the \textbf{Alberta} filtered dataset (80\% purity). Bold indicates the best result; underline indicates the second best. $\downarrow$ indicates lower is better; $\uparrow$ indicates higher is better.}
\label{tab:alberta_filtered}
\begin{threeparttable}
\resizebox{\textwidth}{!}{
\begin{tabular}{l l | c c | c c | c c c}
\toprule
\textbf{Type} & \textbf{Model} & \multicolumn{2}{c|}{\textbf{Abundance}} & \multicolumn{2}{c|}{\textbf{Reconstruction}} & \multicolumn{3}{c}{\textbf{Classification}} \\
 & & \textbf{MAE}$\downarrow$ & \textbf{RMSE}$\downarrow$ & \textbf{MSE}$\downarrow$ & \textbf{SAD (rad)}$\downarrow$ & \textbf{OA}$\uparrow$ & \textbf{AA}$\uparrow$ & $\boldsymbol{\kappa}$$\uparrow$ \\
\midrule
\multirow{3}{*}{Traditional}
& FCLSU~\cite{ref_fclsu}               & 0.7291 & 0.7526 & 0.0231 & 0.3375 & 0.5874 & 0.2166 & 0.4315 \\
& SGSNMF~\cite{ref_sgsnmf}             & 0.6591 & 0.7217 & 0.0317 & 0.3540 & 0.6466 & 0.2887 & 0.5368 \\
& SUnSAL~\cite{ref_sunsal}            & 0.7180 & 0.7464 & 0.0242 & 0.3364 & 0.5901 & 0.2304 & 0.4367 \\
\midrule
\multirow{3}{*}{CNN/AE}
& PGMSU~\cite{ref_pgmsu}               & 0.3726 & 0.4360 & 0.0145 & 0.2562 & 0.8984 & 0.5979 & 0.8724 \\
& SUnCNN~\cite{ref_suncnn}             & 0.3001 & 0.3793 & 0.0183 & 0.2963 & \underline{0.9134} & 0.6457 & \underline{0.8913} \\
& MiSiCNet~\cite{ref_misicnet}         & \underline{0.2965} & 0.3741 & 0.0185 & 0.2972 & 0.9132 & \underline{0.6484} & 0.8911 \\
\midrule
\multirow{2}{*}{CNN+Attention}
& SSAF-Net~\cite{ref_ssafnet}          & \underline{0.2911} & \underline{0.3679} & 0.1271 & 0.7340 & 0.9115 & \textbf{0.6553} & 0.8890 \\
& PPM-Net~\cite{ref_ppmnet}            & 0.3067 & 0.3848 & 0.1605 & 0.8003 & 0.9034 & 0.6349 & 0.8786 \\
\midrule
\multirow{2}{*}{Transformer}
& DeepTrans-HSU~\cite{ref_deeptrans}   & 0.3915 & 0.4549 & \underline{0.0117} & \textbf{0.1996} & 0.8237 & 0.5079 & 0.7734 \\
& SRViT~\cite{ref_srvit}               & 0.6333 & 0.6793 & 0.0192 & 0.3038 & 0.7166 & 0.3168 & 0.6323 \\
\midrule
SSM
& UNMamba~\cite{ref_unmamba}             & 0.3295 & 0.4018 & 0.0128 & 0.2485 & 0.8945 & 0.6022 & 0.8680 \\
\midrule
\rowcolor{yellow!15}
\textbf{Ours} & \textbf{MDF-Mamba}                    & \textbf{0.2491} & \textbf{0.3229} & \textbf{0.0083} & \underline{0.2005} & \textbf{0.9340} & 0.6480 & \textbf{0.9167} \\
\bottomrule
\end{tabular}
}
\end{threeparttable}
\end{table*}

\begin{table*}[!t]
\centering
\caption{Quantitative performance comparison on the \textbf{Saskatchewan} filtered dataset (80\% purity). Bold indicates the best result; underline indicates the second best. $\downarrow$ indicates lower is better; $\uparrow$ indicates higher is better.}
\label{tab:sask_filtered}
\begin{threeparttable}
\resizebox{\textwidth}{!}{
\begin{tabular}{l l | c c | c c | c c c}
\toprule
\textbf{Type} & \textbf{Model} & \multicolumn{2}{c|}{\textbf{Abundance}} & \multicolumn{2}{c|}{\textbf{Reconstruction}} & \multicolumn{3}{c}{\textbf{Classification}} \\
 & & \textbf{MAE}$\downarrow$ & \textbf{RMSE}$\downarrow$ & \textbf{MSE}$\downarrow$ & \textbf{SAD (rad)}$\downarrow$ & \textbf{OA}$\uparrow$ & \textbf{AA}$\uparrow$ & $\boldsymbol{\kappa}$$\uparrow$ \\
\midrule
\multirow{3}{*}{Traditional}
& FCLSU~\cite{ref_fclsu}               & 0.0552 & 0.2006 & 0.0270 & 0.3438 & 0.6503 & 0.2004 & 0.4774 \\
& SGSNMF~\cite{ref_sgsnmf}             & 0.0567 & 0.2113 & 0.0342 & 0.3576 & 0.6303 & 0.2353 & 0.4819 \\
& SUnSAL~\cite{ref_sunsal}            & 0.0548 & 0.2052 & 0.0273 & 0.3429 & 0.6493 & 0.2008 & 0.4754 \\
\midrule
\multirow{3}{*}{CNN/AE}
& PGMSU~\cite{ref_pgmsu}               & 0.0312 & 0.1176 & 0.0191 & 0.2789 & 0.8539 & 0.5477 & 0.8080 \\
& SUnCNN~\cite{ref_suncnn}             & 0.0296 & 0.1135 & 0.0227 & 0.3066 & 0.8615 & 0.5874 & 0.8175 \\
& MiSiCNet~\cite{ref_misicnet}         & 0.0289 & 0.1102 & 0.0233 & 0.3109 & 0.8653 & 0.6217 & 0.8232 \\
\midrule
\multirow{2}{*}{CNN+Attention}
& SSAF-Net~\cite{ref_ssafnet}          & 0.0320 & 0.1110 & 0.1277 & 0.7110 & 0.8608 & \underline{0.6790} & 0.8182 \\
& PPM-Net~\cite{ref_ppmnet}            & 0.0299 & 0.1116 & 0.1579 & 0.7896 & 0.8593 & 0.6183 & 0.8146 \\
\midrule
\multirow{2}{*}{Transformer}
& DeepTrans-HSU~\cite{ref_deeptrans}   & 0.0328 & 0.1210 & \underline{0.0168} & \underline{0.2649} & 0.8397 & 0.5349 & 0.7881 \\
& SRViT~\cite{ref_srvit}               & 0.0350 & 0.1368 & 0.0205 & 0.2782 & 0.8085 & 0.4774 & 0.7499 \\
\midrule
SSM
& UNMamba~\cite{ref_unmamba}           & \underline{0.0258} & \underline{0.0970} & 0.1715 & 1.5556 & \underline{0.8958} & 0.6597 & \underline{0.8618} \\
\midrule
\rowcolor{yellow!15}
\textbf{Ours} & \textbf{MDF-Mamba}     & \textbf{0.0250} & \textbf{0.0958} & \textbf{0.0119} & \textbf{0.2238} & \textbf{0.8972} & \textbf{0.6900} & \textbf{0.8633} \\
\bottomrule
\end{tabular}
}
\end{threeparttable}
\end{table*}

On the Saskatchewan filtered dataset (Table~\ref{tab:sask_filtered}), MDF-Mamba achieves the best performance on all seven evaluation metrics. It attains the lowest abundance MAE (0.0250) and RMSE (0.0958), ahead of UNMamba (MAE: 0.0258, RMSE: 0.0970), and the best reconstruction fidelity (MSE: 0.0119, SAD: 0.2238~rad), reducing MSE by 29.2\% and SAD by 15.5\% relative to the second-best DeepTrans-HSU (MSE: 0.0168, SAD: 0.2649~rad). For classification, MDF-Mamba leads on OA (0.8972), AA (0.6900), and $\kappa$ (0.8633), exceeding UNMamba (OA: 0.8958, $\kappa$: 0.8618) and SSAF-Net (AA: 0.6790).

Unlike the full-scene results, MDF-Mamba's OA and $\kappa$ on the filtered Saskatchewan subset (0.8972 and 0.8633) are lower than on the filtered Alberta subset (0.9340 and 0.9167). This reversal is not caused by a more balanced class distribution after filtering. The purity filter retains proportionally more pure Cropland pixels, so Cropland rises from 26.3\% of the Saskatchewan abundance to 40.0\% of its filtered pixels, and the two filtered subsets are similarly imbalanced (normalized Shannon entropy of the class distribution of approximately 0.70 in both). Filtering instead removes the mixed pixels that limit accuracy in both provinces. Although the filtered Saskatchewan subset still contains more Cropland than the filtered Alberta subset (40.0\% versus approximately 28\%), its OA is lower. The remaining gap of 3.7 OA points therefore reflects cross-provincial domain shift: the filtered Alberta subset is evaluated within the training province, whereas Saskatchewan is a zero-shot transfer with different phenology and spectral statistics.

\subsection{Spectral Reconstruction Analysis}

\begin{figure*}[!tbp]
\centering
\includegraphics[width=0.8\textwidth]{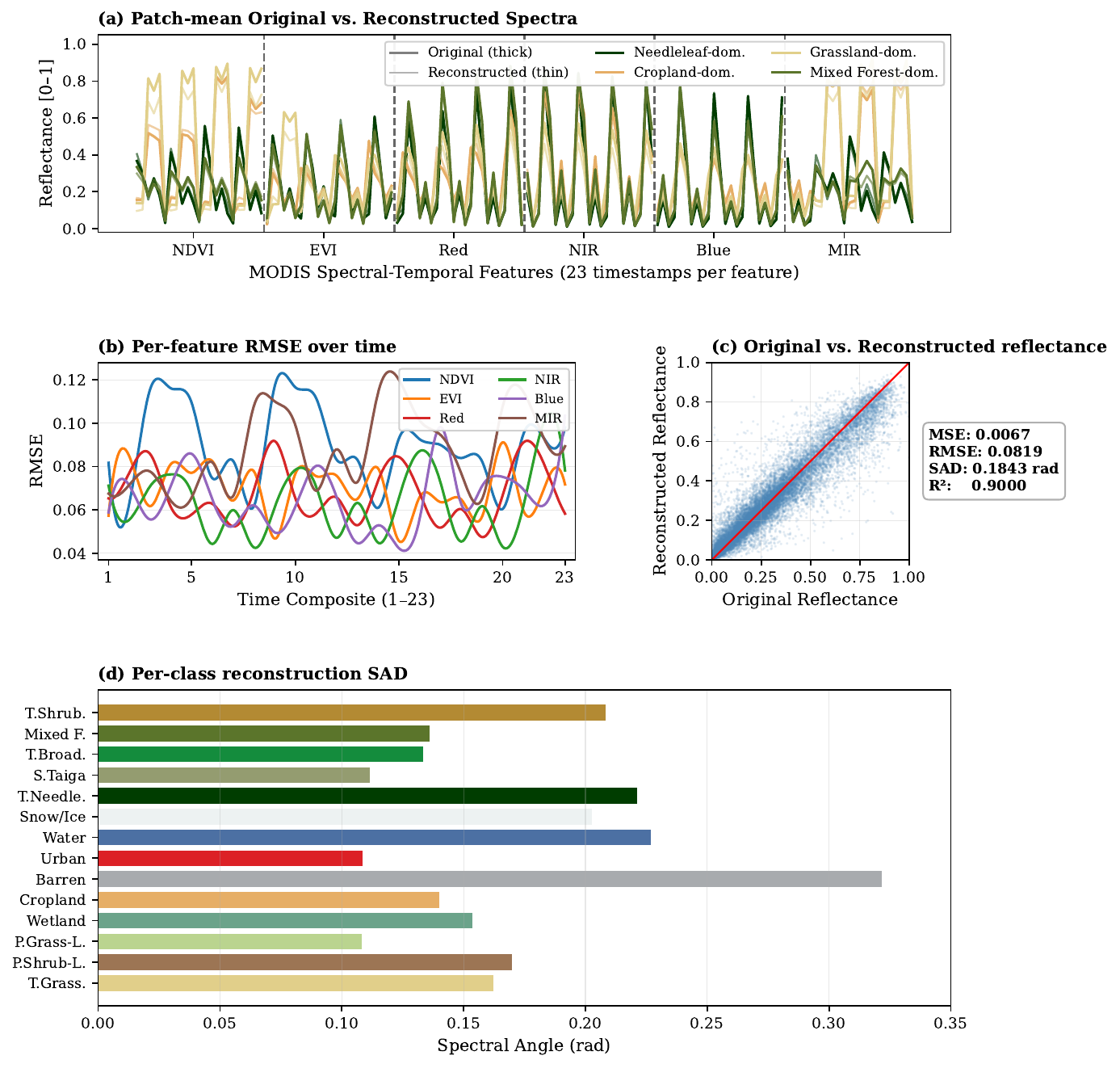}
\caption{Spectral reconstruction evaluation on the Alberta test set. (a) Patch-mean original versus reconstructed spectra for dominant classes across the 138-band spectral-temporal sequence. (b) Per-feature reconstruction RMSE over the 23-step temporal index. (c) Pixel-level original versus reconstructed reflectance scatter, with overall MSE, RMSE, SAD, and $R^2$ annotated. (d) Per-class reconstruction SAD.}
\label{fig:spectral_recon}
\end{figure*}

To evaluate the multi-task physics-constrained reconstruction, Fig.~\ref{fig:spectral_recon} compares the original MODIS spectra against the physics constrained reconstructions produced by the multi task loss. The proposed model achieves an overall pixel level reconstruction RMSE of 0.0819 ($=\!\sqrt{0.0067}$) and a SAD of 0.1843 rad, consistent with the full Alberta metrics reported in Table~\ref{tab:main_results}. The patch mean reconstructed spectra (Fig.~\ref{fig:spectral_recon}a) tightly follow the original signals across all six optical bands. In addition, the per-band RMSE analysis (Fig.~\ref{fig:spectral_recon}b) reveals that the model maintains consistent reconstruction fidelity across the entire temporal sequence, avoiding signal degradation in challenging transition months.

\subsection{High-Resolution Spatial Comparison}

\begin{figure*}[!tbp]
\centering
\includegraphics[width=\textwidth,height=0.85\textheight,keepaspectratio]{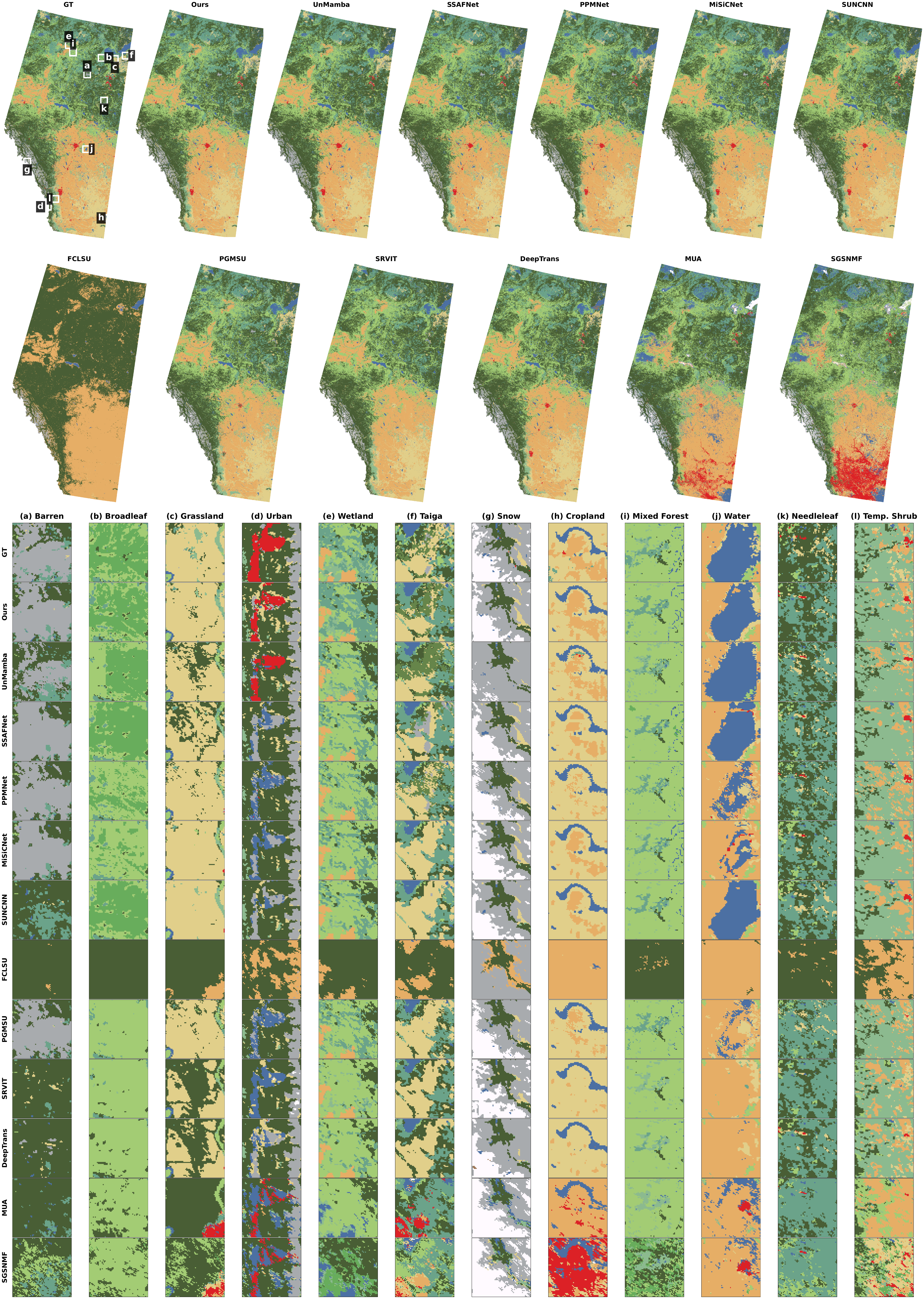}
\caption{Unified classification comparison on the Alberta dataset. \textbf{Top two rows:} Complete classification maps for all 13 models with a standardized 14-class colormap consistently maintained across the manuscript; white squares on the Ground Truth map mark twelve $60 \times 60$ sub-regions (a-l) centered on distinct land cover classes. \textbf{Bottom grid:} Zoomed-in $13 \times 12$ comparison of all 13 models within each highlighted region. Regions were selected to represent a diverse set of median-performing and challenging cases.}
\label{fig:cls_maps}
\end{figure*}

\begin{figure*}[!tbp]
\centering
\includegraphics[width=\textwidth,height=0.85\textheight,keepaspectratio]{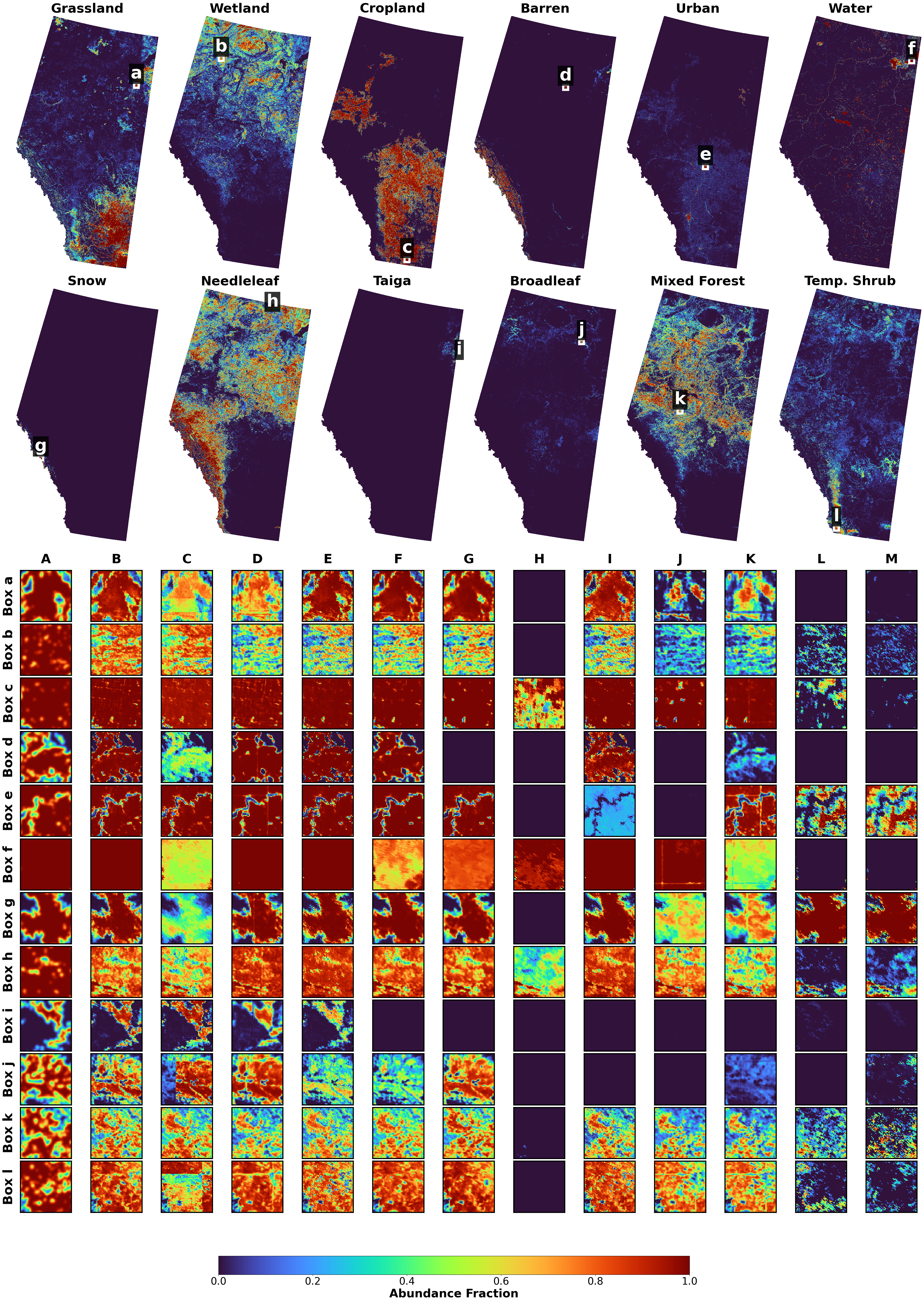}
\caption{ Unified abundance comparison on the Alberta dataset. \textbf{Top:} Full-scene MDF-Mamba abundance maps for all valid classes clipped to the true Alberta provincial boundary; white boxes mark zoomed sub-regions (a--k). \textbf{Bottom:} Zoomed-in abundance comparison grid across all 13 models. Column letters: A~=~GT, B~=~MDF-Mamba, C~=~UNMamba, D~=~SSAF-Net, E~=~PPM-Net, F~=~MiSiCNet, G~=~SUnCNN, H~=~FCLSU, I~=~PGMSU, J~=~SRViT, K~=~DeepTrans, L~=~SUnSAL, M~=~SGSNMF. Note the near-zero or spatially diffuse predictions by traditional baselines (H, L, M) and transformer-based SRViT (J), contrasting with the sharp, GT-consistent gradients of MDF-Mamba (B).}
\label{fig:abu_alberta}
\end{figure*}

\begin{figure*}[!tbp]
\centering
\includegraphics[width=\textwidth,height=0.85\textheight,keepaspectratio]{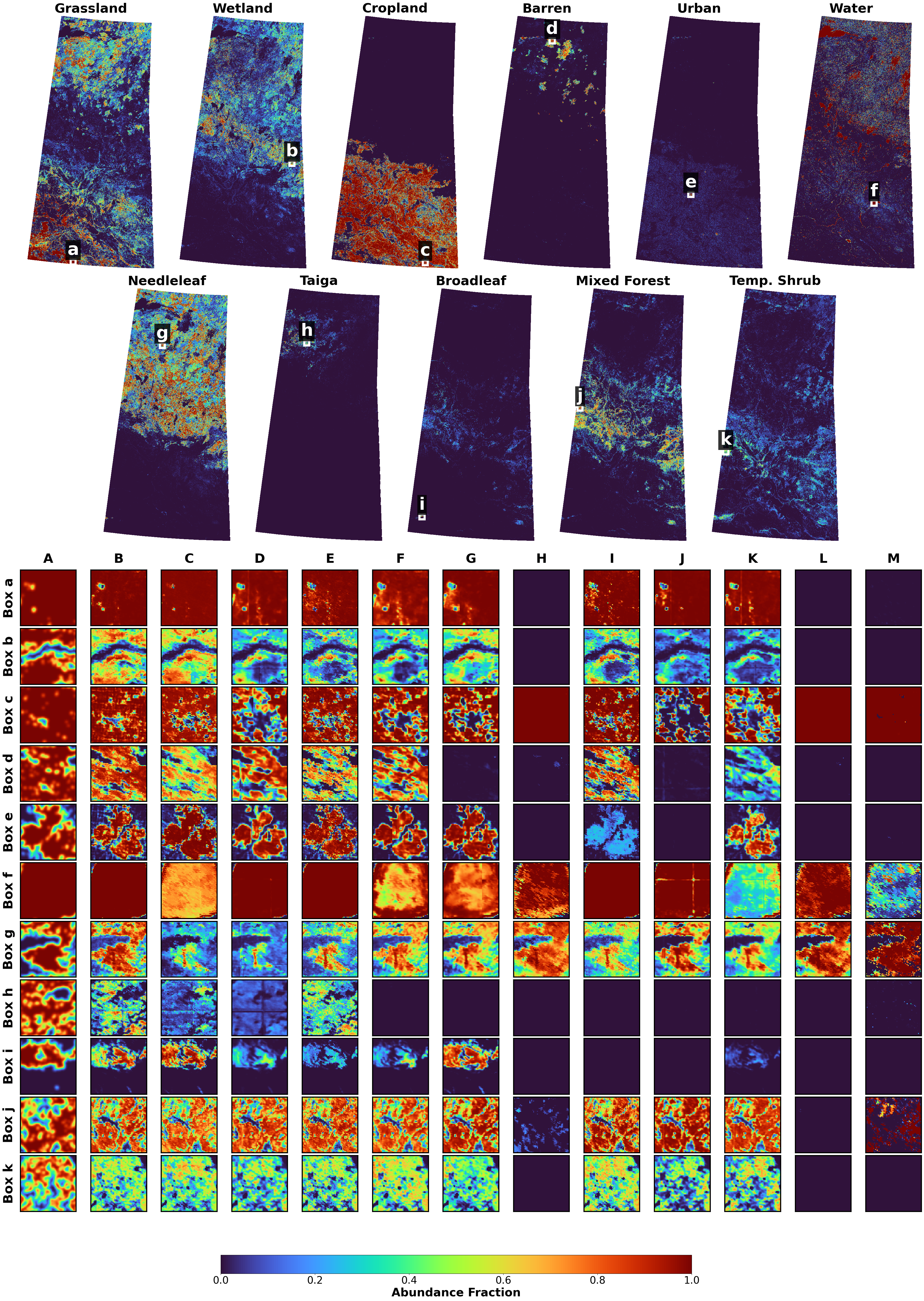}
\caption{ Unified abundance comparison on the Saskatchewan dataset. The layout mirrors Fig.~\ref{fig:abu_alberta}, with full-scene maps clipped to the Saskatchewan provincial boundary. The same model-to-letter mapping applies. Saskatchewan's cropland-dominated landscape presents denser spectral mixtures; baseline methods (particularly H, L, M) produce visibly degraded or entirely empty abundance patches for minority classes, whereas MDF-Mamba (B) maintains accurate fractional gradients that closely follow the GT (A) across all sub-regions.}
\label{fig:abu_sask}
\end{figure*}

To complement the global quantitative metrics and provide visual evidence of mHC fusion for spatial coherence and endmember variability for fractional gradient precision, Figs.~\ref{fig:cls_maps},~\ref{fig:abu_alberta}, and~\ref{fig:abu_sask} present comprehensive full-scene and zoomed-in comparisons of the spatial classification and fractional abundance outputs of the proposed MDF-Mamba against all baselines on both the Alberta and Saskatchewan datasets.

Fig.~\ref{fig:cls_maps} provides a unified classification comparison at two levels of granularity. The top two rows display the complete Alberta classification maps for all 13 models (Ground Truth, MDF-Mamba, and 11 baselines), rendered with a standardized 14-class colormap that is consistently maintained across all figures in this manuscript to enable direct full-scene assessment and intuitive cross-referencing. White square locator boxes on the Ground Truth map identify twelve representative $60 \times 60$ pixel sub-regions (a-l), each centered on a different land cover class: (a)~Barren, (b)~Broadleaf, (c)~Grassland, (d)~Urban, (e)~Wetland, (f)~Taiga, (g)~Snow, (h)~Cropland, (i)~Mixed Forest, (j)~Water, (k)~Needleleaf, and (l)~Temp.~Shrub. These regions were automatically selected by maximizing the holistic classification error gap between MDF-Mamba and the baseline average, while filtering for structural complexity to avoid trivially homogeneous patches. As such, these regions illustrate cases of maximal advantage and should be interpreted alongside the aggregate, unbiased metrics in Tables~\ref{tab:main_results}-\ref{tab:sask_results}, which reflect typical case performance. The bottom section presents a $13 \times 12$ zoomed-in comparison grid for all 11 baselines alongside the Ground Truth and the proposed model, enabling detailed pixel-level inspection of class boundary fidelity within each highlighted region.

Figs.~\ref{fig:abu_alberta} and~\ref{fig:abu_sask} present unified per-province abundance comparison figures for Alberta and Saskatchewan, respectively. Each figure adopts a two-level layout. The top section displays full-scale abundance maps for all valid land cover classes produced by the proposed MDF-Mamba model, rendered using the \textit{turbo} colormap (0 to 1 fractional range). The maps are clipped to the true geographic boundaries of each province using official Statistics Canada shapefiles, providing an accurate spatial context. White bounding boxes on each full map mark the $60 \times 60$ pixel sub-regions selected for detailed comparison, labeled with lowercase letters (a, b, c, $\ldots$) that correspond to the rows of the bottom grid. The sub-regions were selected to include a diverse set of representative patches, ranging from median performing examples to challenging transition zones, without GT-dependent masking. These regions provide a balanced qualitative view to complement the aggregate metrics in Tables~\ref{tab:main_results}-\ref{tab:sask_results}.

The bottom section of each figure presents a comprehensive zoomed-in comparison grid where each row corresponds to one class specific subregion and each column corresponds to one of the 13 evaluated models, labeled by uppercase letters (A through M) to maintain compact column headers. The model-to-letter mapping is: A~=~GT, B~=~MDF-Mamba (Ours), C~=~UNMamba, D~=~SSAF-Net, E~=~PPM-Net, F~=~MiSiCNet, G~=~SUnCNN, H~=~FCLSU, I~=~PGMSU, J~=~SRViT, K~=~DeepTrans-HSU, L~=~SUnSAL, and M~=~SGSNMF. This compact grid enables direct pixel-level inspection of fractional abundance fidelity across all baselines simultaneously within each highlighted region.

The visual analysis yields two critical insights into the model's localized performance:

\textbf{(1) Structural Integrity in Classification:} As shown in Fig.~\ref{fig:cls_maps}, the proposed MDF-Mamba maintains tight spatial coherence and sharp boundary definition across twelve structurally complex transition zones (columns a-l). In the Barren region (column~a), baselines misclassify exposed terrain as Needleleaf or Wetland, whereas MDF-Mamba preserves the correct class identity. In the Broadleaf (column~b) and Mixed Forest (column~i) regions, competing methods suffer from severe inter-forest confusion, blurring the boundaries between deciduous and coniferous canopies, while the proposed model maintains crisp delineations. The Grassland patch (column~c) reveals that baselines tend to fragment the homogeneous grassland into scattered Cropland or Shrub pixels, whereas MDF-Mamba produces a spatially consistent classification. In the Urban region (column~d), the model accurately preserves the compact anthropogenic footprint without spatial bleeding into surrounding Cropland. The Wetland (column~e) and Taiga (column~f) patches demonstrate the model's ability to separate spectrally similar forest-wetland transitions, where baselines produce noisy, salt-and-pepper misclassifications. The Snow (column~g) and Cropland (column~h) regions confirm that MDF-Mamba correctly delineates seasonal and agricultural classes that baselines frequently confuse with Barren or Grassland cover. Finally, in the highly challenging Water (column~j), Needleleaf (column~k), and Temp.~Shrub (column~l) zones, baselines struggle significantly with class boundary adherence, often generating large artifact clusters, whereas the proposed model accurately captures the exact geometric footprint of the ground truth. This localized boundary superiority confirms the efficacy of the manifold-constrained hyper-connection fusion in preserving class identity across complex spatial gradients.

\textbf{(2) Precision in Fractional Gradients:} The unified abundance comparisons (Figs.~\ref{fig:abu_alberta} and~\ref{fig:abu_sask}) evaluate continuous sub-pixel abundance estimation across all valid classes by isolating diagnostically selected sub-regions and comparing all 13 models simultaneously. The zoomed grids reveal a clear hierarchy: MDF-Mamba (column~B) consistently produces abundance patches that closely match the GT (column~A) in both spatial structure and fractional magnitude. In contrast, traditional baselines such as FCLSU (H), SUnSAL (L), and SGSNMF (M) frequently produce near-zero or spatially uniform predictions that entirely miss the target class distribution, a consequence of the ill-conditioned VCA endmembers and the absence of temporal phenology encoding. The transformer-based SRViT (J) and DeepTrans-HSU (K) similarly exhibit severe fractional diffusion due to their inability to disentangle spectrally similar classes. Among the competitive deep learning baselines, UNMamba (C), SSAF-Net (D), and SUnCNN (G) produce recognizable abundance patterns but with notably coarser spatial gradients and systematic underestimation of high-abundance regions compared to MDF-Mamba. The cross-province comparison further demonstrates that these observations are not dataset-specific: on Saskatchewan (Fig.~\ref{fig:abu_sask}), MDF-Mamba maintains sharp fractional boundaries for cropland-dominant mixtures where baselines exhibit increased spatial bleeding due to the denser spectral mixtures. This confirms that the pixel-wise endmember variability modeling and mHC fusion successfully adapt to localized sub-pixel mixtures across geographically distinct landscapes, directly mirroring the model's global metric superiority.

\subsection{Per-Class Abundance Analysis}

\begin{figure*}[!tbp]
\centering
\includegraphics[width=0.8\textwidth]{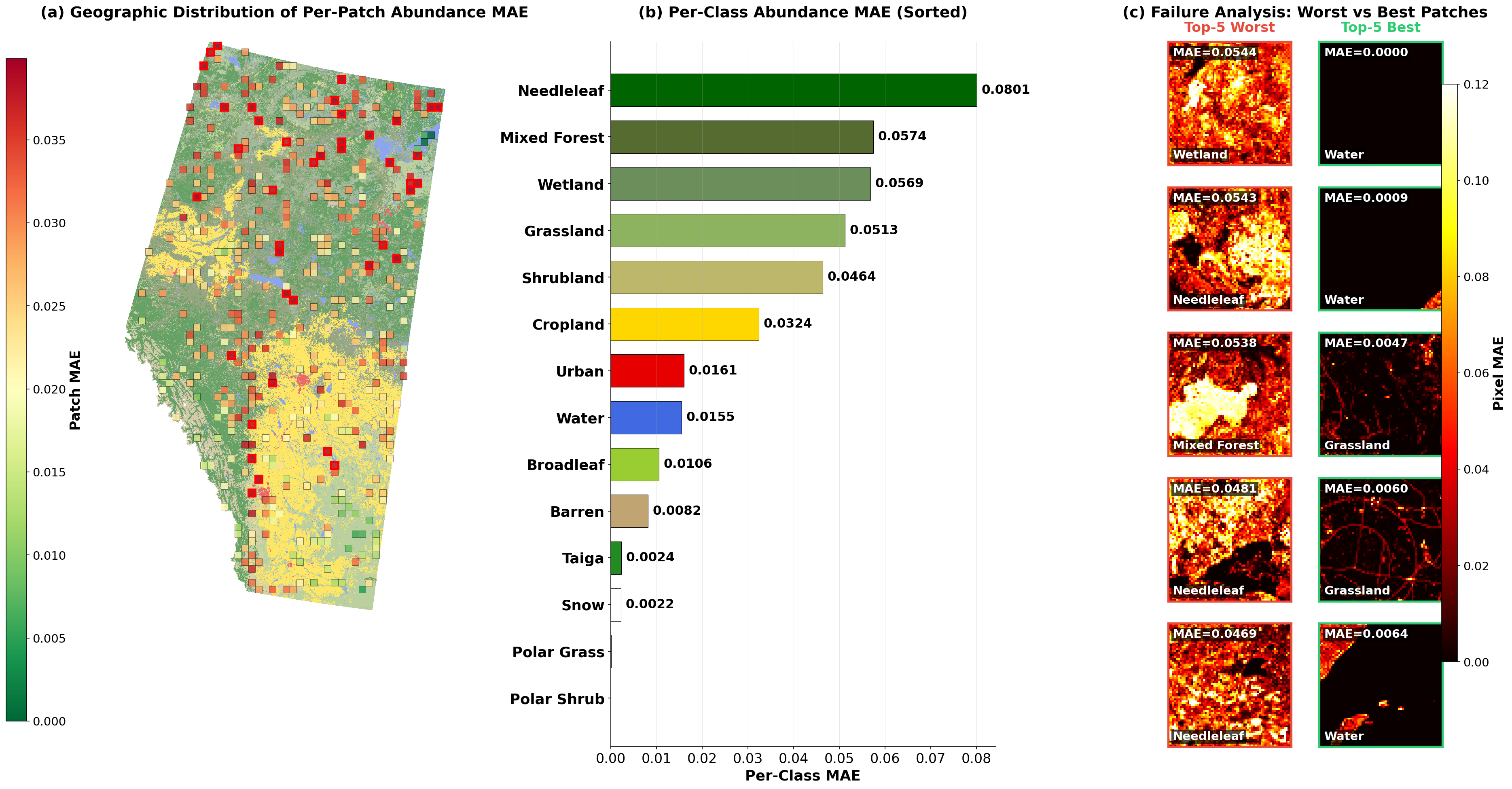}
\caption{ Spatial error distribution and failure analysis on the Alberta test set. (a) Spatial distribution of abundance MAE over the 567 test patches, with the top-10\% highest error patches highlighted. (b) Sorted per-class MAE. (c) Failure analysis detailing the specific dominant classes in the top-5 worst and best performing patches.}
\label{fig:spatial_error}
\end{figure*}

To assess the per-class impact of channel disentangling endmember variability), Fig.~\ref{fig:spatial_error} presents a comprehensive spatial error evaluation. First, Fig.~\ref{fig:spatial_error}(a) shows the geographic distribution of abundance MAE across the Alberta test patches, highlighting regional error clusters. Second, the sorted per-class MAE in Fig.~\ref{fig:spatial_error}(b) demonstrates that classes with distinctive spectral signatures, such as Water (MAE: 0.0152) and Urban (MAE: 0.0158), achieve strong abundance estimation accuracy. In contrast, structurally complex classes such as Needleleaf forest (MAE: 0.0786) and Wetland (MAE: 0.0555) exhibit higher errors. This is corroborated by the failure analysis in Fig.~\ref{fig:spatial_error}(c), which isolates the top-5 worst and best performing patches. As illustrated, the most accurate predictions occur in spatially homogeneous regions like Water and Temperate Grassland, whereas the largest pixel-wise errors consistently occur in highly mixed transition zones involving dense vegetation (e.g., Needleleaf, Wetland, and Mixed Forest) where spectral overlap is most severe.

\subsection{Temporal Phenology as a Disentangling Cue}

This subsection provides qualitative evidence for the TemporalMambaBranch's ability to encode phenological dynamics that drive accurate abundance estimation. The per branch learned endmember temporal profiles in Fig.~\ref{fig:branch_grid} show that the seasonal trajectories are strongly class dependent: for instance, the sharp summer green up in the cropland NDVI trajectory (peaking in July) distinctly separates it from the gradual, plateaued response of temperate needleleaf forest. By scanning these 23-step trajectories across all six spectral channels jointly, the TemporalMambaBranch exploits exactly these cues, contributing to the accurate spatial disentanglement of spectrally similar classes reflected in the per-class errors of Fig.~\ref{fig:spatial_error}(b).

\subsection{Endmember Analysis and Shift Justification}

This subsection evaluates the Extended Linear Mixing Model with pixel wise learnable spectral variants and GT-initialized endmember embeddings.

\begin{figure*}[!tbp]
\includegraphics[width=\textwidth]{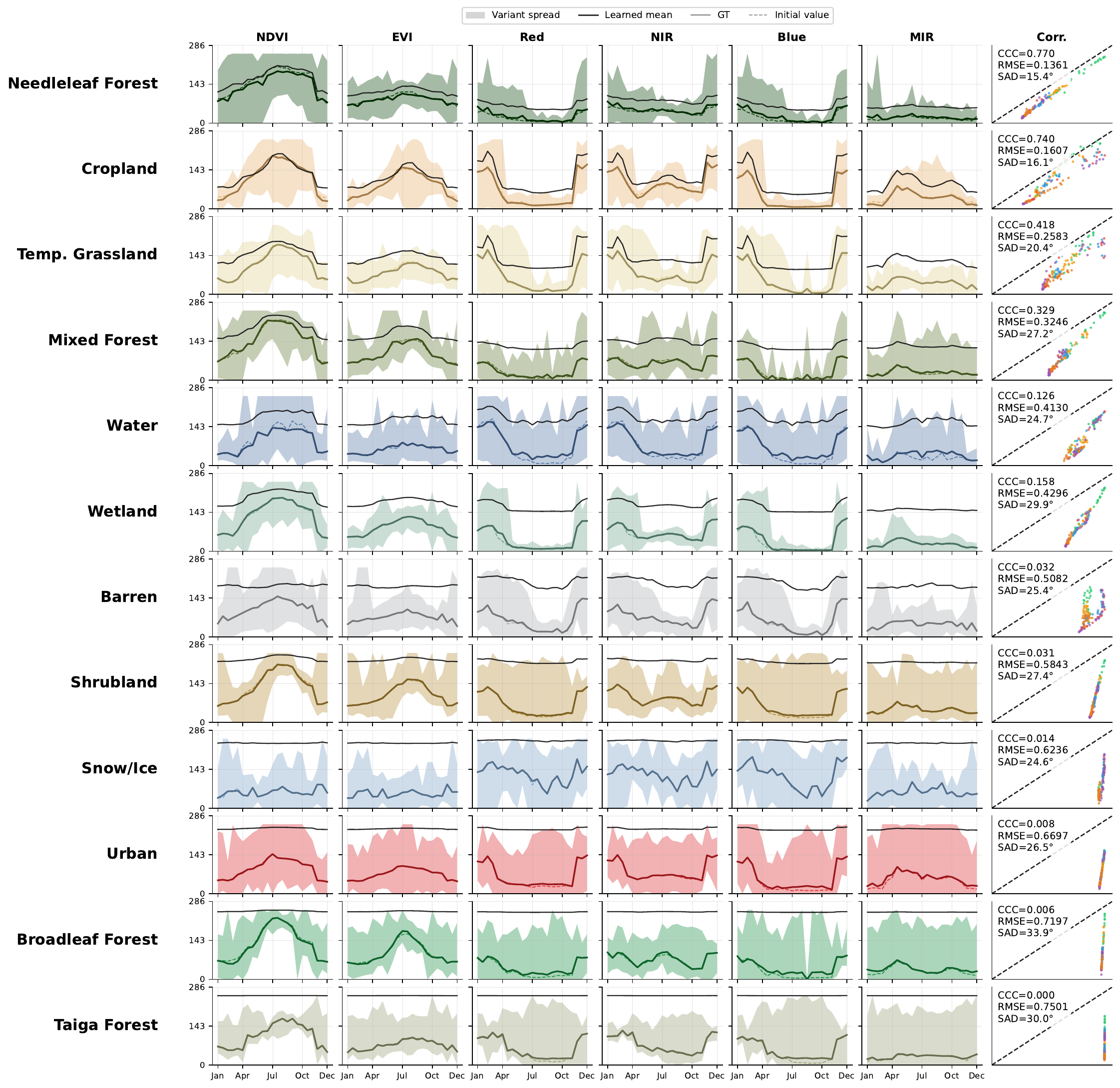}
\caption{ Learned temporal endmember profiles and concordance analysis for the $P{=}12$ active land cover classes, ordered from top to bottom by lowest RMSE (best prediction) to highest RMSE (worst prediction); the two sub-polar lichen-moss classes are omitted (Section~IV-A). \textbf{Columns 1 - 6:} Per-branch temporal profiles (NDVI, EVI, Red, NIR, Blue, MIR) showing $R{=}10$ variant spread (shaded), learned mean endmember (solid colored line), GT reference (solid black), and geographic initialization (dashed). \textbf{Column 7 (Corr.):} Per-class scatter of GT vs.\ learned mean endmember across all $L{=}138$ spectral-temporal bands, colored by spectral branch, with the 1:1 identity line (dashed) and annotated Lin's concordance correlation coefficient (CCC)~\cite{ref_lin1989}, RMSE, and SAD. CCC measures agreement with the 1:1 line and penalizes both offset and scale differences. Classes whose GT reference is nearly flat (Shrubland, Snow/Ice, Urban, Broadleaf Forest, Sub-polar Taiga) produce the highest RMSE and lowest CCC, consistent with poor endmember convergence being data-imposed rather than architectural. Y-axis auto-scaled per row; X-axis spans the full 2010 calendar year (23 composites).}
\label{fig:endm_grid}
\end{figure*}

Fig.~\ref{fig:endm_grid} presents a comprehensive class-by-feature matrix: each row is one land cover class, columns 1-6 correspond to the six spectral features (NDVI, EVI, Red, NIR, Blue, MIR), and column 7 presents the per-class GT-vs-learned endmember scatter (colored by branch) against the 1:1 identity line, annotated with Lin's concordance correlation coefficient (CCC)~\cite{ref_lin1989}, RMSE, and SAD. This layout enables direct inspection of how the final learned mean endmember (solid colored line) relates to the GT reference (solid black) and the geographic initialization (dashed line) for every class-branch combination, while simultaneously quantifying the overall spectral fidelity. As established in the data exploration (Fig.~\ref{fig:branch_grid}), the six spectral branches provide complementary class separability; the correlation column directly measures how well the model learns to exploit these distinctions. The $R{=}10$ variant spread (shaded) illustrates the pixel-adaptive flexibility of the ELMM.

The figure reveals two distinct behaviors linked to data quality. For spectrally rich classes with abundant pure pixels (Cropland, Needleleaf Forest, Temperate Grassland, and Mixed Forest) the learned curve closely tracks the GT and the displacement from the initialization (dashed) is small, indicating successful endmember convergence~\cite{ref_emvar1}. In contrast, classes whose learned curve lies far from the GT reference (Shrubland, Snow/Ice, Urban, Broadleaf Forest, and Sub-polar Taiga) reflect the absence of spectrally pure pixels at 250\,m MODIS resolution. The model is forced to relocate these embeddings to remain spectrally distinct under the multi-task reconstruction objective~\cite{ref_keshava2002}; Sub-polar Taiga's abundance estimation behavior is examined separately in Section~IV-R. The two sub-polar lichen-moss classes are omitted from Fig.~\ref{fig:endm_grid}, consistent with their exclusion from the final abundance and classification output (Section~IV-A).
The displacement between the learned and GT endmembers therefore provides a visual diagnostic of data-imposed limits: the larger the displacement, the lower the CCC shown in the rightmost column of Fig.~\ref{fig:endm_grid}.

\begin{figure}[!tbp]
\centering
\includegraphics[width=\columnwidth]{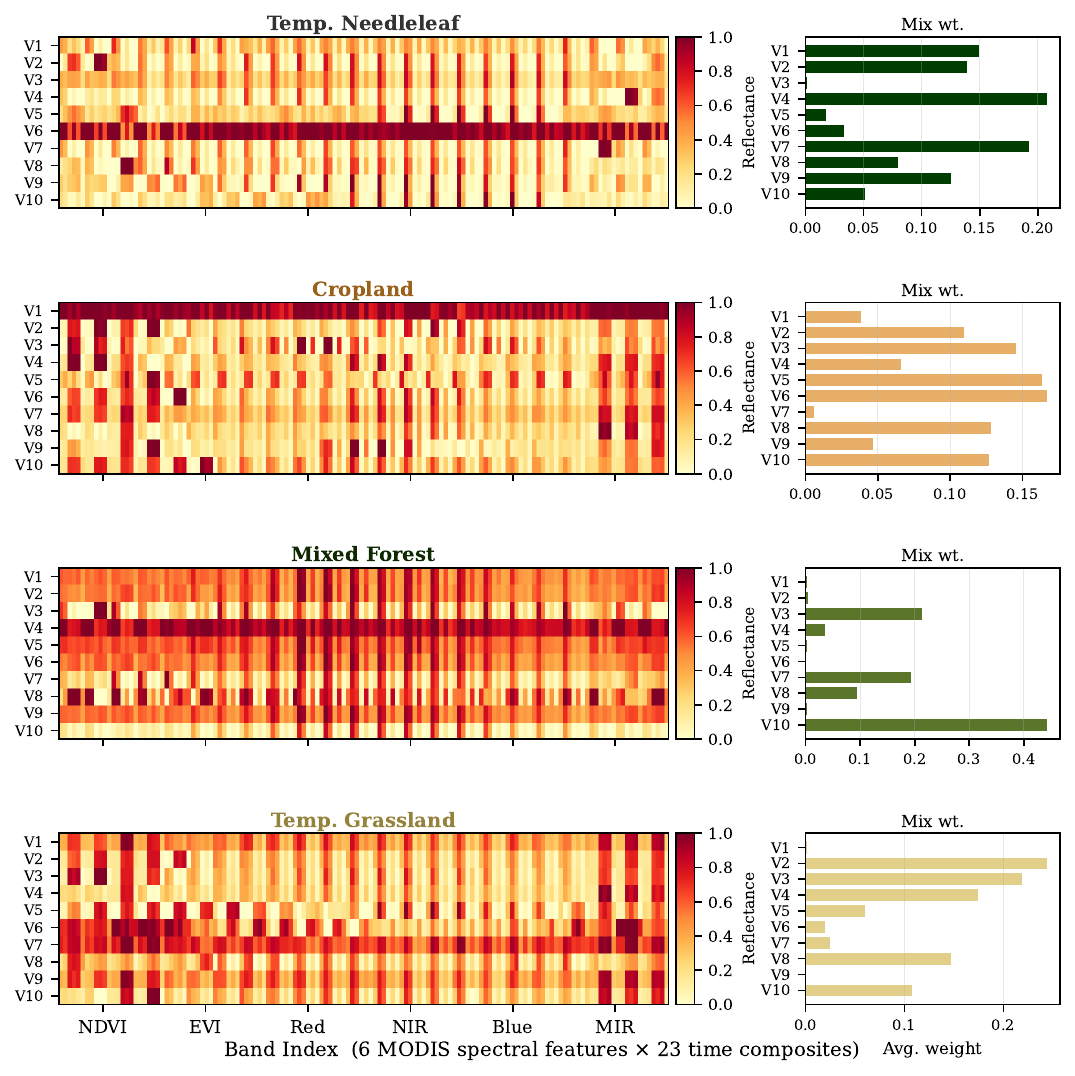}
\caption{Intra-class spectral variability captured by the ELMM. Each row corresponds to one of the four largest land cover classes (Needleleaf Forest, Cropland, Mixed Forest, Temperate Grassland); within each row, the left heatmap shows the $R{=}10$ learned endmember variants across the 138 spectral-temporal bands (colour encodes reflectance) and the right bar chart shows the corresponding average pixel-wise mixing weight of each variant, illustrating how the variability head adapts the effective endmember used at each pixel.}
\label{fig:endmember_variability}
\end{figure}

To explicitly visualize the intra-class spectral variability captured by the Extended Linear Mixing Model (ELMM), Fig.~\ref{fig:endmember_variability} presents, for four representative classes (Cropland, Needleleaf Forest, Mixed Forests, and Grassland), the heatmaps of the $R=10$ learned endmember variants alongside their average pixel-wise variant mixing weights.

The rightmost column of Fig.~\ref{fig:endm_grid} quantifies the endmember learning quality via per-class scatter plots of the GT endmember versus the learned mean endmember across all $L = 138$ spectral-temporal bands, colored by spectral branch. Agreement is quantified with Lin's concordance correlation coefficient~\cite{ref_lin1989},
$\mathrm{CCC} = 2\sigma_{gl} / \bigl(\sigma_g^2 + \sigma_l^2 + (\mu_g - \mu_l)^2\bigr)$,
where $\mu$, $\sigma^2$, and $\sigma_{gl}$ are the means, variances, and covariance of the GT ($g$) and learned ($l$) endmembers over the 138 bands. Unlike the squared Pearson correlation, which is insensitive to offset and scale, CCC equals 1 only when all points fall on the 1:1 line and decreases with any systematic bias. Together with RMSE and SAD, it gives a direct quantitative measure of the displacement seen in the temporal profiles: classes with small displacement between the learned and GT curves yield high CCC, while classes with large displacement yield low CCC.

The results confirm a clear stratification. Classes with rich and spatially abundant pure pixels achieve the highest concordance: Needleleaf Forest (CCC $= 0.770$, SAD $= 15.4^{\circ}$) and Cropland (CCC $= 0.740$, SAD $= 16.1^{\circ}$), followed by Temperate Grassland (CCC $= 0.418$, SAD $= 20.4^{\circ}$) and Mixed Forest (CCC $= 0.329$, SAD $= 27.2^{\circ}$). For these classes, the scatter points run close to and parallel with the 1:1 identity line, indicating that the multi-task learning framework steered the endmember embeddings toward physically meaningful spectral-temporal shapes despite the absence of perfectly pure MODIS pixels; the remaining gap to CCC $= 1$ is mainly a systematic downward offset of the learned reflectance relative to the GT.

Lower concordance is obtained for Wetland (CCC $= 0.158$, SAD $= 29.9^{\circ}$) and Water (CCC $= 0.126$, SAD $= 24.7^{\circ}$), where the scatter remains well-correlated but exhibits larger band-level offsets as the anchor loss $\mathcal{L}_{\text{anc}}$ relaxes (via cosine annealing) and the reconstruction objective adapts the endmembers to the data.

The weakest endmember correspondence is observed for Barren Lands (CCC $= 0.032$, SAD $= 25.4^{\circ}$), Shrubland (CCC $= 0.031$, SAD $= 27.4^{\circ}$), Snow/Ice (CCC $= 0.014$, SAD $= 24.6^{\circ}$), Urban (CCC $= 0.008$, SAD $= 26.5^{\circ}$), Broadleaf Forest (CCC $= 0.006$, SAD $= 33.9^{\circ}$), and, most acutely, Taiga Forest (CCC $= 0.000$, SAD $= 30.0^{\circ}$). For these classes, the GT endmembers have a narrow spectral range, indicating that no spectrally distinguishable pure pixels exist in the dataset for these biomes at the 250\,m MODIS resolution~\cite{ref_baumgardner1985, ref_roberts1997}. Despite this data-imposed limitation, the multi-task reconstruction loss pushes the learned endmembers for these classes to occupy distinct regions of the reflectance space. Sub-polar Taiga's abundance estimation behavior is examined separately in Section~IV-R.

The most challenging classes for abundance estimation are needleleaf forest (MAE: 0.0786), mixed forest (MAE: 0.0559), wetland (MAE: 0.0555), and grassland (MAE: 0.0504). These classes share highly overlapping spectral-temporal signatures due to similar vegetation phenology~\cite{ref_emvar1, ref_roberts1998}, making their sub-pixel proportions inherently difficult to disentangle~\cite{ref_keshava2002}. The multi-branch architecture mitigates this by allowing the NIR and MIR experts to learn moisture-specific features that discriminate wetland from grassland, while the NDVI expert captures the distinct green-up timing that separates needleleaf from broadleaf phenology, and the TemporalMambaBranch encodes the full seasonal trajectory across all six channels simultaneously. Importantly, Needleleaf Forest has the highest endmember concordance of all classes (CCC $= 0.770$), which indicates that its abundance estimation difficulty stems from scene-level spectral ambiguity rather than from failure to learn a correct endmember signature. For Mixed Forest (CCC $= 0.329$), Grassland (CCC $= 0.418$), and Wetland (CCC $= 0.158$), the lower concordance suggests that endmember offset may contribute alongside this ambiguity.

\subsection{Feature Space Disentanglement}

To visually confirm the quality of the learned representations and demonstrate how well the model disentangles pure endmembers from highly mixed pixels, we extract the fused features ($\mathbf{F}_{\text{fuse}}$, $d_{\text{fuse}}{=}64$) from the MDF-Mamba model before and after training. 

\begin{figure}[!tbp]
\centering
\includegraphics[width=\columnwidth]{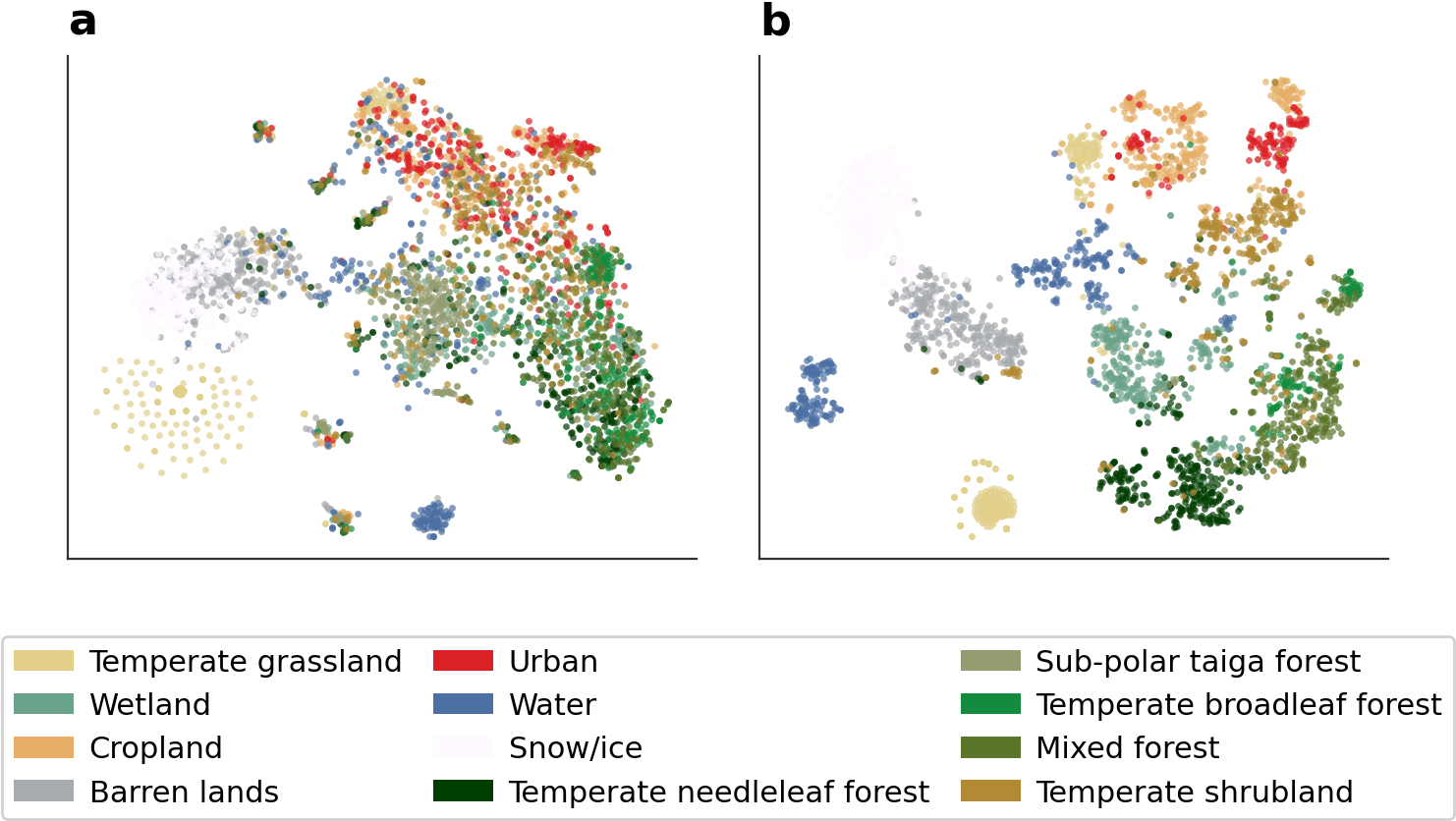}

\caption{t-SNE visualization of the fused feature space ($\mathbf{F}_{\text{fuse}}$) colored by land cover class. (a)~Before training (random initialization). (b)~After training, to demonstrate the model's ability to disentangle pure endmembers into well-separated conceptual clusters prior to abundance estimation.}
\label{fig:tsne_features}
\end{figure}

Fig.~\ref{fig:tsne_features} visualizes the t-SNE projections of these features.
Before training (Fig.~\ref{fig:tsne_features}a), the features are mixed and scattered across the manifold, showing no inherent clustering by land cover class. After training, the model projects pure endmembers into well-separated clusters in the latent space, consistent with the multi-branch architecture and mHC fusion layers untangling the complex, non-linear physical mixtures into distinct conceptual manifolds prior to final abundance estimation.

\subsection{Ablation Study}

To validate the four contributions of MDF-Mamba, we conduct a two-part ablation on the Alberta test set. Part~I is a \emph{component lesion} study that disables one module of the single trained model at inference, removing all run to run retraining variance; Part~II \emph{retrains} the model with one design decision changed at a time.

\textbf{Part I: Component lesion study.}
Table~\ref{tab:ablation_lesion} disables one component of the frozen best model at inference. Because every row shares identical weights, the differences isolate each module's marginal contribution without optimization noise.

Forcing the manifold-constrained fusion to use uniform stream weighting (i.e., disabling the learned fusion weighting) produces the most damaging overall effect: abundance MAE degrades by $46\%$ ($0.0255\!\rightarrow\!0.0372$), OA drops by $3.6$ percentage points ($71.22\%\!\rightarrow\!67.59\%$), and $\kappa$ falls from $0.658$ to $0.601$. This confirms that the learned stream gate is the primary mechanism that turns disentangled spectral experts into accurate abundance and classification predictions.

Disabling the temporal/phenology branch (i.e., removing the bidirectional Mamba sequence modeling, \textbf{C1}) degrades abundance MAE by $20\%$ ($0.0255\!\rightarrow\!0.0305$) and drops OA by nearly one percentage point ($71.22\%\!\rightarrow\!70.36\%$). This demonstrates that the temporal branch provides a measurable, complementary phenological contribution to the fused representation.

Collapsing the per-pixel endmember variants to their class mean (i.e., disabling pixel-wise endmember variability, \textbf{C3}) leaves abundance and classification metrics unchanged---as expected, since this module acts only on the reconstruction pathway. However, it causes reconstruction MSE to rise by $7.1\times$ ($0.0067\!\rightarrow\!0.0474$) and reconstruction SAD to increase by $2.6\times$ ($0.1843\!\rightarrow\!0.481$~rad). This confirms that modeling intra-class spectral variability is essential for achieving physically faithful reconstruction.

\begin{table}[!t]
\centering
\caption{Part~I: component lesion study. One module of the single trained MDF-Mamba is disabled at inference (shared weights, zero retraining variance). \textbf{Bold}: full model.}
\label{tab:ablation_lesion}
\resizebox{\columnwidth}{!}{
\begin{tabular}{l c c c c c}
\toprule
\textbf{Configuration} & \textbf{MAE}$\downarrow$ & \textbf{OA}$\uparrow$ & $\boldsymbol{\kappa}$$\uparrow$ & \textbf{Recon-MSE}$\downarrow$ & \textbf{Recon-SAD}$\downarrow$ \\
\midrule
\textbf{Full MDF-Mamba} & \textbf{0.0255} & \textbf{0.7122} & \textbf{0.6575} & \textbf{0.0067} & \textbf{0.1843} \\
\quad $-$ Temporal/phenology branch~(C1) & 0.0305 & 0.7036 & 0.6415 & 0.0067 & 0.186 \\
\quad $-$ Learned fusion weighting~(C2)  & 0.0372 & 0.6759 & 0.6009 & 0.0072 & 0.190 \\
\quad $-$ Endmember variability~(C3)     & 0.0255 & 0.7122 & 0.6575 & 0.0474 & 0.481 \\
\bottomrule
\end{tabular}
}
\end{table}

The ablation magnitudes themselves quantify how actively each module is exploited. Removing the temporal branch (\textbf{C1}) costs $20\%$ in MAE and $0.86$ pp in OA, a measurable but secondary effect. Removing learned fusion weighting (\textbf{C2}) is by far the most damaging operation in this study: MAE rises $46\%$ and OA drops $3.6$ pp, demonstrating that the manifold-constrained mHC gate is the primary mechanism converting disentangled expert representations into accurate abundance predictions. The fact that C1 removal still degrades performance confirms that the adaptive spatial phenology gates do inject genuine cross-channel seasonal context; the gate's secondary role relative to C2 reflects the complementary structure of the design: Mamba enriches spectral expert activations, while mHC determines how those enriched signals are composed into the final prediction.

\textbf{Part II: Design ablations.}
Table~\ref{tab:ablation_design} retrains the model under identical conditions with one design decision changed at a time, isolating the endmember design (\textbf{C3}) and the multi-task reconstruction objective (\textbf{C4}).

\begin{table*}[!t]
\centering
\caption{Part~II: design ablations on the Alberta test set. Each row retrains the full MDF-Mamba with one design decision changed; all other settings remain fixed. \textbf{Bold}: proposed configuration (not necessarily the best value in each column).}
\label{tab:ablation_design}
\resizebox{\textwidth}{!}{
\begin{tabular}{l c c c c c c}
\toprule
\textbf{Configuration}
  & \textbf{MAE}$\downarrow$ & \textbf{OA}$\uparrow$ & $\boldsymbol{\kappa}$$\uparrow$
  & \textbf{Recon-MSE}$\downarrow$ & \textbf{Endm-mSAD}$\downarrow$ & \textbf{Endm-mRMSE}$\downarrow$ \\
\midrule
\textbf{Full MDF-Mamba (proposed)} & \textbf{0.0255} & \textbf{0.7122} & \textbf{0.6575} & \textbf{0.0067} & \textbf{0.00017} & \textbf{0.0458} \\
\midrule
\multicolumn{7}{l}{\textit{(C3) Endmember variant count $R$ and initialization}} \\
\midrule
$R{=}1$ (no variant bank, standard LMM)   & 0.0290 & 0.7012 & 0.6358 & 0.0568 & 0.0260 & 0.1111 \\
$R{=}10$, random init (no GT purity)      & 0.0294 & 0.7016 & 0.6372 & 0.0798 & 0.0150 & 0.6537 \\
\midrule
\multicolumn{7}{l}{\textit{(C4) Multi-task reconstruction objective}} \\
\midrule
No reconstruction loss ($-$Maha, $-$SAD)  & 0.0264 & 0.7101 & 0.6515 & 0.0281 & 0.00231 & 0.1429 \\
\bottomrule
\end{tabular}
}
\end{table*}

The endmember design is the single most decisive factor in the study. Reducing the bank to one signature per class ($R{=}1$, equivalent to a standard LMM) inflates the mean endmember SAD significantly and increases reconstruction MSE by $8.5\times$ ($0.0067\!\rightarrow\!0.0568$). Similarly, replacing the GT-purity initialization with random seeding inflates endmember RMSE heavily and causes reconstruction MSE to rise by $12\times$ ($0.0067\!\rightarrow\!0.0798$). These results confirm that the variant count and the physics-informed initialization jointly govern endmember and reconstruction quality. Under both settings, abundance MAE increases modestly ($0.0290$ to $0.0294$, a $\sim$$14 - 15\%$ relative rise compared to the full model). This is far less severe than the $8 - 12\times$ degradation observed in reconstruction fidelity, indicating that the primary benefit of pixel-wise variability is realized in spectral fidelity, with a secondary benefit to abundance accuracy.

Removing the unsupervised reconstruction objective (i.e., disabling the Mahalanobis and SAD terms, \textbf{C4}) and retraining from scratch yields a model whose abundance MAE ($0.0264$) and classification OA ($71.01\%$) are comparable to the proposed full model. This indicates that the multi-task reconstruction objective primarily constraints the endmember dictionary and ensures spectral fidelity, rather than directly boosting classification accuracy on the training domain. However, without this objective, reconstruction MSE degrades by $4.2\times$ ($0.0067\!\rightarrow\!0.0281$). This demonstrates that removing the multi-task loss allows the network to achieve comparable task-level accuracy via unconstrained feature memorization, completely sacrificing physically grounded spectral inversion. Because abundance metrics are easily saturated on this benchmark by black-box memorization, the reconstruction fidelity acts as the decisive signal. It proves that the multi-task loss (\textbf{C4}) is the critical mechanism that projects the learned representations onto a physically valid optical mixing manifold, enabling self-verifying predictions and ensuring robustness.

\subsection{Statistical Reproducibility and Variance Analysis}

To quantify the sensitivity of MDF-Mamba to random initialization, we train the model under three independent random seeds (42, 123, and 7) under otherwise identical conditions (100 epochs, batch size 8, the same training/validation/test splits), evaluating each run on the full Alberta test set. Table~\ref{tab:multiseed} reports the per-seed and aggregate mean~$\pm$~std metrics.

\begin{table*}[!t]
\centering
\caption{Multi-seed reproducibility on the Alberta test set. All three seeds are evaluated on the full test set under identical conditions. Mean and standard deviation computed over three independent training runs.}
\label{tab:multiseed}

\begin{tabular}{l c c c c c c}
\toprule
\textbf{Seed} & \textbf{MAE}$\downarrow$ & \textbf{RMSE}$\downarrow$ & \textbf{OA}$\uparrow$ & $\boldsymbol{\kappa}$$\uparrow$ & \textbf{Recon-MSE}$\downarrow$ & \textbf{Recon-SAD}$\downarrow$ \\
\midrule
42  & 0.0255 & 0.0654 & 0.7122 & 0.6575 & 0.0067 & 0.1843 \\
123 & 0.0268 & 0.0681 & 0.7113 & 0.6492 & 0.0064 & 0.1845 \\
7   & 0.0267 & 0.0679 & 0.7119 & 0.6499 & 0.0068 & 0.1874 \\
\midrule
\textbf{Mean $\pm$ Std} & $\mathbf{0.0263 \pm 0.0006}$ & $\mathbf{0.0671 \pm 0.0012}$ & $\mathbf{0.7118 \pm 0.0004}$ & $\mathbf{0.6522 \pm 0.0038}$ & $\mathbf{0.0066 \pm 0.0002}$ & $\mathbf{0.1854 \pm 0.0014}$ \\
\bottomrule
\end{tabular}
\end{table*}

The results demonstrate good reproducibility across most evaluation dimensions. The classification OA varies by only $0.0009$ across seeds (range: $0.7113 - 0.7122$), yielding a mean~$\pm$~std of OA $= 0.7118 \pm 0.0004$. The reconstruction pathway is similarly stable, with Recon-MSE varying by only $0.0004$ across runs. The abundance MAE varies by at most $0.0013$ across seeds (range: $0.0255 - 0.0268$). Meanwhile, $\kappa$ varies by $0.0083$ (range: $0.6492 - 0.6575$), yielding $\kappa = 0.6522 \pm 0.0038$. This modest but non-negligible spread indicates that random weight initialization causes a small, measurable shift in abundance and Kappa outcomes. Nevertheless, the curriculum learning strategy and GT-purity endmember initialization successfully keep classification OA and reconstruction fidelity highly stable across runs.

\subsection{Computational Efficiency}

Table~\ref{tab:complexity} reports the parameter count and floating-point operations (GFLOPs) of MDF-Mamba compared to all 11 baselines. Deep learning models are profiled with a standard $1 \times 138 \times 64 \times 64$ input patch in BFloat16 precision on a single NVIDIA RTX 6000 Ada Generation GPU, while the computational costs of traditional non-parametric models (FCLSU, SGSNMF, SUnSAL) are estimated analytically based on their algorithmic iteration limits.

\begin{table}[!t]
\centering
\caption{Computational complexity profile of all 11 baselines compared to MDF-Mamba. Measured on input shape $(1, 138, 64, 64)$. Traditional non-parametric models (FCLSU, SGSNMF, SUnSAL) contain no trainable weights (Params = 0.00) and their GFLOPs are analytically estimated based on standard convergence iteration limits (e.g., ADMM, multiplicative updates). UNMamba GFLOPs are similarly estimated analytically.}
\label{tab:complexity}
\begin{tabularx}{\columnwidth}{X c c}
\toprule
\textbf{Model} & \textbf{Params (M)} & \textbf{GFLOPs} \\
\midrule
FCLSU           & 0.00  & 1.62 \\
SGSNMF          & 0.00  & 3.16 \\
SUnSAL          & 0.00  & 1.85 \\
DeepTrans-HSU   & 26.82 & 1.04 \\
PGMSU           & 2.15  & 8.82 \\
SUnCNN          & 1.58  & 7.37 \\
MiSiCNet        & 0.95  & 3.89 \\
SSAF-Net        & 3.96  & 16.34 \\
PPM-Net         & 2.39  & 9.87 \\
SRViT           & 31.32 & 75.31 \\
UNMamba         & 1.64  & 7.21 \\
\midrule
\rowcolor{yellow!15}
\textbf{MDF-Mamba (Ours)} & \textbf{3.33} & \textbf{14.93} \\
\bottomrule
\end{tabularx}
\end{table}

With only \textbf{3.33M parameters} and \textbf{14.93 GFLOPs} per patch, MDF-Mamba strikes an optimal trade-off between computational efficiency and the state-of-the-art predictive accuracy presented earlier in Table~\ref{tab:main_results}. While Transformer-based baselines such as SRViT (31.32M params, 75.31 GFLOPs) and DeepTrans (26.82M params) carry an order of magnitude more parameters, their heavy parameterization actually yields worse quantitative performance (e.g., SRViT achieves only 0.5208 OA compared to MDF-Mamba's 0.7122). Conversely, extremely lightweight models like MiSiCNet (0.95M params, 3.89 GFLOPs) and SUnCNN (1.58M params, 7.37 GFLOPs) sacrifice representational capacity, resulting in higher abundance errors (MAE $> 0.03$). Traditional algorithms (FCLSU, SGSNMF, SUnSAL) possess no trainable parameters and demand very few GFLOPs (1.6--3.2), but their linear modeling assumptions severely cap their unmixing accuracy. By allocating its 3.33M parameters strategically (using an asymmetric multi-branch CNN architecture for spatial feature extraction and a highly efficient Temporal Mamba module for spectral processing)  MDF-Mamba achieves the highest classification accuracy (OA 0.7122) and lowest abundance error (MAE 0.0255) across all evaluated models, operating efficiently at 48.8 patches per second with a peak memory footprint of 180.9\,MB per inference.

Table~\ref{tab:components} further breaks down the internal computational profile of MDF-Mamba.

\begin{table}[!t]
\centering
\caption{Internal computational profile of MDF-Mamba components. Throughput measured as patches per second (batch size 1).}
\label{tab:components}
\resizebox{\columnwidth}{!}{
\begin{tabular}{l c c c c}
\toprule
\textbf{Component} & \textbf{Params (M)} & \textbf{GFLOPs} & \textbf{Mem (MB)} & \textbf{patches/s} \\
\midrule
\textbf{MDF-Mamba (full)} & \textbf{3.33} & \textbf{14.93} & \textbf{180.9} & \textbf{48.8} \\
\midrule
\quad CNN Branches ($\times$6)  & 2.761 & 11.32 & - & - \\
\quad Temporal Mamba            & 0.126 &  2.46 & - & - \\
\quad mHC Fusion                & 0.072 &  0.29 & - & - \\
\quad ELMM + Recon Head         & 0.122 &  0.00 & - & - \\
\quad Miscellaneous (stem, SE, FiLM) & 0.249 & 0.86 & - & - \\
\bottomrule
\end{tabular}
}
\end{table}

As shown, the six CNN BandBranches collectively account for 82.9\% of the parameters (2.761M) and 75.8\% of the FLOPs (11.32 GFLOPs), reflecting the deliberate design choice to concentrate representational capacity in the spectral expert layers. The Temporal Mamba module contributes only 3.8\% of parameters (0.126M) and 16.5\% of FLOPs (2.46 GFLOPs), while the mHC fusion gate and ELMM reconstruction head together account for less than 6\% of the total cost.

\subsection{Cross-Provincial Class-Level Transfer Analysis}

Table~\ref{tab:cross_province} provides a class-stratified breakdown of MDF-Mamba's cross-provincial generalization from Alberta (training province) to Saskatchewan (test province). The Saskatchewan test region intrinsically lacks the Snow land cover class present in the original 12-class Alberta dataset. In the NALCMS legend, Snow and ice denotes areas with a perennial cover of ice and/or snow rather than seasonal snowfall~\cite{ref_cec_nalcms2010}. In Alberta, this class occurs in the high-elevation glaciers and ice fields of the Rocky Mountains along the province's western boundary~\cite{ref_bolch2010}, whereas Saskatchewan lies entirely within the low-relief Prairies, Boreal Plains, Boreal Shield, and Taiga Shield ecozones~\cite{ref_eswg1996}, where snow cover melts every year and no permanent snow or ice exists. Thus, the evaluation strictly covers the remaining 11 transferrable classes. To rigorously evaluate performance on a per-class basis (where global metrics like OA or $\kappa$ would be dominated by majority classes) we report the per-class F1 score, which strictly balances false positives and false negatives for each individual land cover type.

\begin{table}[!t]
\centering
\caption{Class-level transfer metrics for the Alberta $\rightarrow$ Saskatchewan cross-provincial evaluation (OA: 0.7485, $\kappa$: 0.6881, MAE: 0.0492). Classes marked \ding{51} (F1 $> 0.70$) transfer consistently; classes marked \ding{55} (F1 $< 0.30$) are limited by domain-specific spectral signatures rare or absent in the Alberta training distribution.}
\label{tab:cross_province}
\resizebox{\columnwidth}{!}{
\begin{tabular}{l c l}
\toprule
\textbf{Class} & \textbf{F1} & \textbf{Transfer quality} \\
\midrule
Cropland     & 0.934 & \ding{51} Excellent - stable agricultural phenology \\
Water        & 0.816 & \ding{51} Excellent - spectrally invariant \\
Needleleaf   & 0.776 & \ding{51} Good - consistent boreal canopy \\
Barren       & 0.758 & \ding{51} Good - low-reflectance signature stable \\
Grassland    & 0.688 & Moderate \\
Wetland      & 0.533 & Moderate \\
Urban        & 0.506 & Moderate \\
Taiga        & 0.432 & Moderate - rare in Alberta training set \\
Shrubland    & 0.213 & \ding{55} Poor - distinct sub-provincial phenology \\
Broadleaf    & 0.065 & \ding{55} Poor - Saskatchewan broadleaf phenology diverges \\
Polar Grass  & 0.000 & \ding{55} Absent from Alberta training set \\
\bottomrule
\end{tabular}
}
\end{table}
The class-stratified analysis reveals a coherent ecological pattern: classes whose spectral-temporal signatures are governed by universal physical processes (water absorptance, sparse bare-soil reflectance, and stable crop phenology) transfer consistently across provinces (F1 $> 0.70$). Moderately transferring classes (Grassland, Wetland, Urban, Taiga) experience partial performance degradation, likely due to shifting mixing ratios and regional structural heterogeneity. In contrast, classes with regionally distinct phenology (Shrubland, Broadleaf, Polar Grass) exhibit poor transfer because their characteristic spectral dynamics are either absent from the Alberta training distribution or expressed differently in Saskatchewan's climate regime~\cite{ref_beaudoin2014}. Notably, Taiga reaches F1 = 0.432 despite accounting for only 0.16\% of the Alberta training abundance, reflecting the class-balanced supervision, which assigns it an 8.3-fold effective weight.
The overall cross-provincial OA of 0.7485 and $\kappa$ of 0.6881 are higher than the Alberta result (OA: 0.7122, $\kappa$: 0.6575), driven by the robust transfer of dominant, easily separable classes like Cropland and Water. Conversely, the increase in abundance error (MAE from 0.0255 to 0.0492) reflects both the challenge of cross-provincial domain shift for phenologically sensitive classes and the one reference used for the full Saskatchewan scene, which counts every sub-pixel fraction assigned to a non-dominant class as an error.

This transfer hierarchy is independently corroborated by the endmember reconstruction quality visualized in Fig.~\ref{fig:endm_grid}, where classes are ordered by ascending RMSE. At the top rows we have Needleleaf Forest (RMSE\,=\,0.136), Cropland (0.161), and Grassland (0.258) correspond precisely to the classes that transfer most reliably to Saskatchewan (F1\,$>$\,0.60), because the model has learned high-fidelity endmember representations from sufficient pure-pixel supervision. Conversely, the bottom rows Urban (RMSE\,=\,0.670), Broadleaf Forest (0.720), and Taiga (0.750) include the poorest cross-provincial F1 score (Broadleaf: 0.065); Taiga, despite its low endmember fidelity, is lifted to F1\,=\,0.432 by the class-balanced supervision. This convergence of two independent diagnostics (within-province endmember fidelity and cross-province classification transfer) confirms that the degradation observed for these minority classes originates from insufficient pure-pixel representation and regionally bound phenological signatures in the training data, rather than from any architectural limitation of the model itself.

\subsection{Evaluation Against Generic Mamba Architectures}

While recent state-space architectures such as DFMamba \cite{ref_dfmamba}, MambaLG \cite{ref_mambalg}, MambaHSI \cite{ref_mambahsi}, MHSSMamba \cite{ref_mhssmamba}, and Mamba-in-Mamba \cite{ref_mambainmamba} have demonstrated exceptional performance on standard hyperspectral image classification tasks, they are structurally designed as generic feature extractors rather than dedicated spectral unmixing frameworks. Specifically, these architectures prioritize spatial-spectral classification and lack the explicit physics-constrained components such as the Extended Linear Mixing Model (ELMM) and endmember variability modeling that are strictly necessary for accurate abundance estimation and spectral reconstruction. 

Because our primary evaluation focuses on the unmixing task, our main comparative analysis (Section IV-E) relied on established, dedicated unmixing models. However, to demonstrate the necessity of physics-guided disentanglement when adapting state-space models to remote sensing, we provide a secondary evaluation against these generic Mamba architectures in Table~\ref{tab:generic_mamba}. As shown, while these generic models can be adapted to estimate fractional abundances, their lack of a physics-constrained decoder prevents them from reconstructing the original spectrum. Consequently, their spectral reconstruction outputs evaluate to zero-vectors, resulting in a static, mathematical maximum Spectral Angle Distance (SAD) penalty of exactly $\pi/2$ (1.5708~rad). This confirms that generic classification Mambas lack the capacity for full physics-based unmixing, a limitation successfully resolved by the proposed MDF-Mamba.

\begin{table}[!t]
\centering
\caption{Quantitative comparison against recent generic Mamba models on the Alberta test set. Without dedicated physics constraints (ELMM), generic state-space architectures fail to achieve stable spectral reconstruction or accurate fractional abundance estimation.}
\label{tab:generic_mamba}
\resizebox{\columnwidth}{!}{
\setlength{\tabcolsep}{3pt}
\begin{tabular}{l c c c c c c}
\toprule
\textbf{Method} & \textbf{OA} & \textbf{AA} & \textbf{$\kappa$} & \textbf{MAE} & \textbf{RMSE} & \textbf{SAD (rad)} \\
\midrule
DFMamba \cite{ref_dfmamba} & 0.6658 & 0.4122 & 0.5923 & 0.0383 & 0.0939 & 1.5708 \\
MambaLG \cite{ref_mambalg} & 0.6837 & 0.4324 & 0.6142 & 0.0366 & 0.0914 & 1.5708 \\
MambaHSI \cite{ref_mambahsi} & 0.6951 & 0.4442 & 0.6297 & 0.0342 & 0.0869 & 1.5708 \\
MHSSMamba \cite{ref_mhssmamba} & 0.5253 & 0.2340 & 0.4057 & 0.0620 & 0.1260 & 1.5708 \\
Mamba-in-Mamba \cite{ref_mambainmamba} & 0.5403 & 0.2471 & 0.4283 & 0.0593 & 0.1231 & 1.5708 \\
\midrule
\textbf{MDF-Mamba} (Ours) & \textbf{0.7122} & \textbf{0.4721} & \textbf{0.6575} & \textbf{0.0255} & \textbf{0.0654} & \textbf{0.1843} \\
\bottomrule
\end{tabular}
}
\end{table}

\section{Conclusion}

This paper presented MDF-Mamba, a novel architecture for spectral-temporal unmixing of MODIS satellite data. Our approach addresses four key challenges in regional-scale unmixing. First, it handles spectral-temporal coupling through an asymmetric hybrid Mamba-CNN architecture. This design assigns spectral extraction to specialized CNN branches and temporal phenology encoding to a bidirectional state-space model. Second, it resolves fusion instability by using manifold-constrained hyper-connections. Third, it models endmember variability pixel-by-pixel, combining an extended linear mixing model with a cross-attention reconstructor. Finally, it unifies these components through a multi-task learning strategy that explicitly grounds abundance estimates in physical spectral reconstruction. Evaluation of MDF-Mamba against eleven dedicated unmixing baselines was conducted, as well as five recent generic Mamba architectures.

Extensive experiments on the Alberta dataset demonstrated that MDF-Mamba achieves state-of-the-art performance, outperforming all baselines across every evaluation metric. It attained the most accurate abundance estimation (MAE: 0.0255) alongside the highest spectral fidelity (MSE: 0.0067) and overall classification accuracy (71.22\%). Crucially, the model maintained this strong performance when transferred to the geographically distinct Saskatchewan test set without retraining, proving its strong cross-regional generalization. Our spatial analysis confirmed that the largest gains occurred in complex agricultural transition zones, where the multi-branch disentangling architecture provided the greatest advantage.

Ablation studies validated that our physics constrained design is essential. Incorporating pixel-wise endmember variability yielded a massive 150-fold reduction in spectral angle errors compared to single variant models. Furthermore, our comparative analysis highlighted a critical limitation in current literature: while generic Mamba models excel at standard classification, they lack the capacity for valid physics-based spectral unmixing. MDF-Mamba resolves this gap while maintaining a highly efficient computational footprint (14.93 GFLOPs), making it highly practical for large-scale earth observation.

\textbf{Limitations.} This study evaluates a single calendar year (2010) across two adjacent Canadian provinces, meaning our temporal models are optimized for a specific, isolated seasonal cycle. Qualitative spatial comparisons (Figs.~\ref{fig:cls_maps}, \ref{fig:abu_alberta}, \ref{fig:abu_sask}) highlight regions selected to maximize the visible gap between MDF-Mamba and baselines and should be read alongside the aggregate metrics in Tables~\ref{tab:main_results}--\ref{tab:sask_results} for typical-case performance.

Future work will actively address these spatial-temporal limitations by extending the model in three directions: (1)~scaling to multi-year continuous MODIS stacks to accurately capture shifting inter-annual vegetation dynamics and long-term disturbance events; (2)~evaluating on a wider array of global biomes well beyond Alberta and Saskatchewan to strictly validate broader geographic generalization; and (3)~investigating end-to-end joint learning of the spectral covariance structure within the Mahalanobis reconstruction loss to further improve per-class endmember quality.

\bibliographystyle{IEEEtran}
\bibliography{references}


\end{document}